\documentclass{article}

\PassOptionsToPackage{numbers, compress}{natbib}

\usepackage[preprint]{neurips_2026}

\usepackage[utf8]{inputenc} 
\usepackage[T1]{fontenc}    
\usepackage{hyperref}       
\hypersetup{
    colorlinks=true,
    citecolor=magenta,
}
\usepackage{url}            
\usepackage{booktabs}       
\usepackage{amsfonts}       
\usepackage{nicefrac}       
\usepackage{microtype}      
\usepackage{xcolor}         
\usepackage{pifont}
\usepackage{amssymb}
\usepackage{amsmath}
\usepackage{booktabs}
\usepackage{multirow}
\usepackage[table]{xcolor} 
\usepackage{adjustbox}
\usepackage{amsmath} 
\usepackage{graphicx}
\usepackage{pifont}
\usepackage{caption}
\usepackage{listings}
\definecolor{opensrcHdr}{HTML}{C8DDFB}
\definecolor{opensrcRow}{HTML}{EAF0FB} 
\definecolor{propriHdr}{HTML}{E0D4F5} 
\definecolor{propriRow}{HTML}{F2EDF9} 
\definecolor{spatialHdr}{HTML}{C8D5F0}
\definecolor{spatialRow}{HTML}{E8EDF8} 
\definecolor{rankBest}{HTML}{BDF7C8}
\definecolor{rankSecond}{HTML}{D7FBDD}
\definecolor{rankThird}{HTML}{ECFDEE}
\definecolor{appendixBoxBg}{HTML}{F6F8FB}
\definecolor{appendixPromptBg}{HTML}{EEF5FF}
\definecolor{appendixBoxRule}{HTML}{5B6F95}
\newcommand{\bestcell}[1]{\cellcolor{rankBest}\textbf{#1}}
\newcommand{\secondcell}[1]{\cellcolor{rankSecond}#1}
\newcommand{\thirdcell}[1]{\cellcolor{rankThird}#1}

\DeclareRobustCommand{\topone}{%
  \begingroup\setlength{\fboxsep}{1pt}\colorbox{green!25}{\strut top-1}\endgroup%
}
\DeclareRobustCommand{\toptwo}{%
  \begingroup\setlength{\fboxsep}{1pt}\colorbox{green!16}{\strut top-2}\endgroup%
}
\DeclareRobustCommand{\topthree}{%
  \begingroup\setlength{\fboxsep}{1pt}\colorbox{green!8}{\strut top-3}\endgroup%
}

\newcommand{\modelicon}[1]{\includegraphics[height=1em]{#1}}
\usepackage{wrapfig}

\usepackage[most]{tcolorbox}
\usepackage{tabularx}
\usepackage{enumitem}
\usepackage{listings}

\definecolor{promptBg}{RGB}{248,250,252}
\definecolor{promptRule}{RGB}{203,213,225}
\definecolor{promptTitle}{RGB}{30,41,59}
\definecolor{promptCodeBg}{RGB}{246,248,250}

\lstdefinestyle{allocentricbox}{
  basicstyle=\scriptsize\ttfamily,
  backgroundcolor=\color{appendixBoxBg},
  frame=single,
  rulecolor=\color{appendixBoxRule},
  breaklines=true,
  columns=fullflexible,
  keepspaces=true,
  showstringspaces=false,
  xleftmargin=0.5em,
  xrightmargin=0.5em,
  framexleftmargin=0.5em,
  framexrightmargin=0.5em,
  aboveskip=0.6em,
  belowskip=0.6em
}

\newtcolorbox{promptbox}[2][]{
  enhanced,
  breakable,
  colback=promptBg,
  colframe=promptRule,
  coltitle=white,
  colbacktitle=promptTitle,
  title=\textbf{#2},
  fonttitle=\small,
  fontupper=\small,
  boxrule=0.6pt,
  arc=2mm,
  left=1.2mm,
  right=1.2mm,
  top=1mm,
  bottom=1mm,
  before skip=0.6em,
  after skip=0.8em,
  #1
}

\lstdefinestyle{promptjson}{
  basicstyle=\ttfamily\footnotesize,
  backgroundcolor=\color{promptCodeBg},
  frame=single,
  rulecolor=\color{promptRule},
  breaklines=true,
  columns=fullflexible,
  keepspaces=true,
  showstringspaces=false,
  xleftmargin=0.5em,
  xrightmargin=0.5em
}

\title{AlloSpatial: Agentic Harness Framework for Spatial Reasoning in Foundation Models}

\author{%
  \textbf{Shouwei Ruan}$^{1}$ \quad
  \textbf{Bin Wang}$^{2}$ \quad
  \textbf{Zhenyu Wu}$^{1}$ \quad
  \textbf{Qihui Zhu}$^{1}$ \quad
  \textbf{Yuxiang Zhang}$^{2}$ \quad
  \textbf{Jingzhi Li}$^{3}$ \\[0.5em]
  \textbf{Yubin Wang}$^{2}$\thanks{Project leader.} \quad
  \textbf{Xingxing Wei}$^{1}$\thanks{Corresponding author.} \\[1em]
  $^{1}$ Institute of Artificial Intelligence, Beihang University \\
  $^{2}$ Huawei Noah's Ark Lab \\
  $^{3}$ University of Science and Technology Beijing
}

\begin{document}

\maketitle

\begin{abstract}
Multimodal Foundation Models (MFMs) have made substantial progress, yet remain fragile in spatial reasoning over the physical world. A key bottleneck lies in their inability to transform local egocentric observations into a global allocentric spatial representation. To address this, we propose \textbf{AlloSpatial}, an agentic framework for allocentric spatial cognition in foundation models. AlloSpatial introduces \textbf{World2Mind}, a plug-and-play cognitive mapping sandbox that converts egocentric observations into structured allocentric priors, including Allocentric-Spatial Trees and route maps that support querying object topology, geometric relations, passability, and trajectories. To utilize these priors reliably under noisy reconstruction and ambiguous visual evidence, AlloSpatial introduces a \textbf{Spatial Reasoning Harness} for tool-use judgment, modality-decoupled cue collection, and geometry-semantic arbitration. We further internalize this process in Qwen3-VL through cold-start reinforcement learning with a harness-gated trajectory-level reward. Experiments on VSI-Bench and MindCube show that AlloSpatial improves proprietary models by 5\%--18\% in a training-free setting, while ASTs alone support strong spatial reasoning even when visual inputs are removed. The trained AlloSpatial agents further outperform larger general-purpose models and competitive spatial baselines, suggesting that structured allocentric representations, active tool use, and verifiable reasoning offer a promising route toward spatially capable foundation models.
Our code: \href{https://github.com/Heathcliff-saku/AlloSpatial}{https://github.com/Heathcliff-saku/AlloSpatial}
\end{abstract}

\section{Introduction}

Multimodal Foundation Models (MFMs)~\cite{bai2025qwen3,openai2023gpt4v_systemcard,singh2025openai,gemini3pro,su2025thinking} have made substantial progress in cross-modal understanding and reasoning. Yet their ability to reason about 3D space in the physical world remains fragile~\cite{yang2025thinking,wang2025mindcube,lin2025mmsi,huang2025si,ramakrishnan2024does}. A central limitation is that current MFMs largely operate over local, transient, and egocentric observations, lacking a mechanism to transform partial perceptual evidence into global, persistent, and queryable mental representations that reliably bridge semantic understanding and geometric relations~\cite{wang2025mindcube,su2025thinking,yang2025cambrian}.

\begin{figure*}[t]
    \centering
    \includegraphics[width=0.95\textwidth]{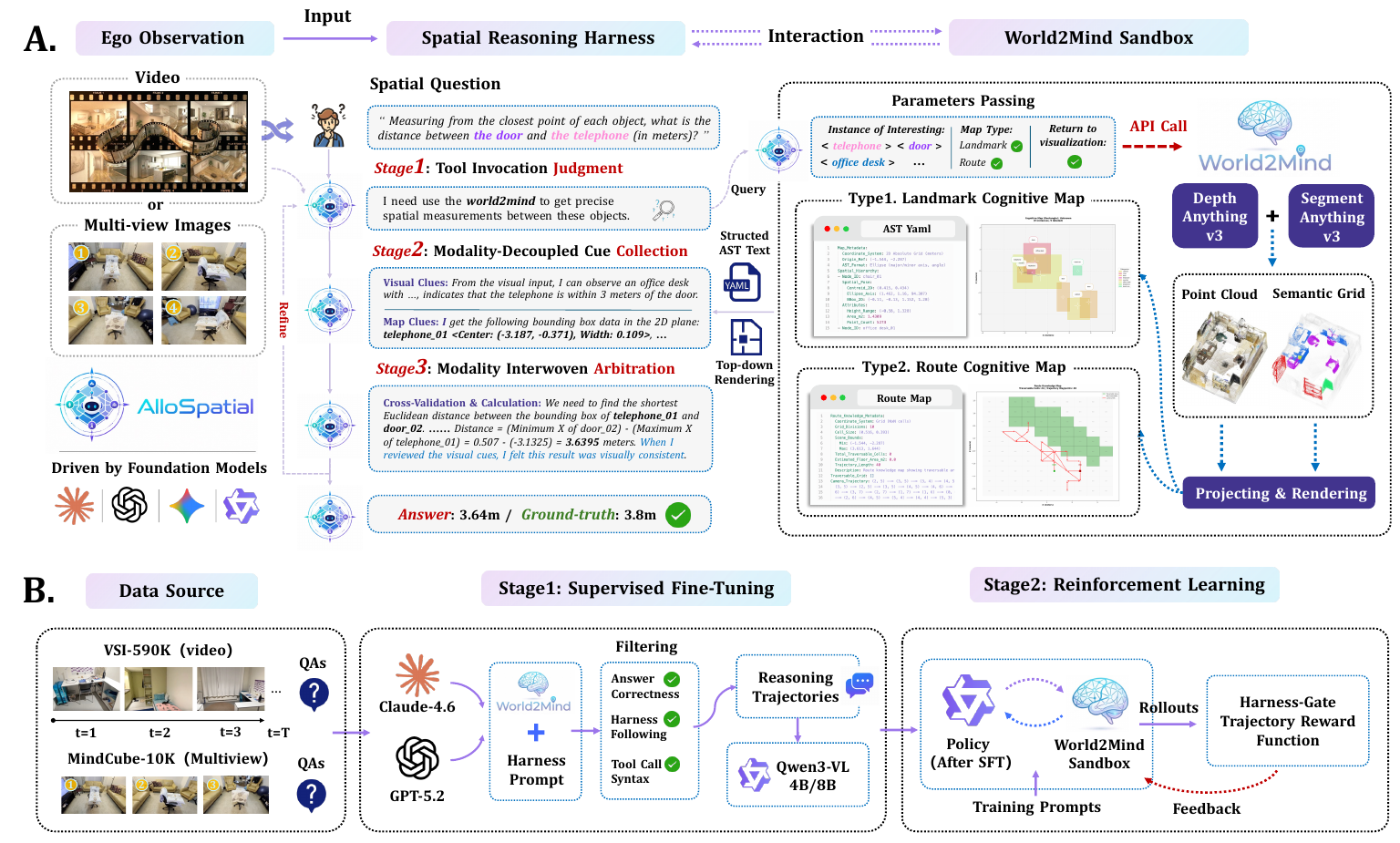}
   \caption{\textbf{AlloSpatial inference and training pipeline.}
\textbf{A}, At inference time, AlloSpatial takes egocentric videos or multi-view observations and follows a three-stage Spatial Reasoning Harness to invoke World2Mind, acquire allocentric spatial knowledge, and arbitrate evidence before answering. \textbf{B}, To internalize this harness, we distill and filter high-quality harness-following trajectories from proprietary models for supervised cold start, and further optimize the policy with live World2Mind interaction and a Harness-Gated Trajectory Reward.}
\vspace{-0.5cm}
    \label{fig:teaser}
\end{figure*}

Existing approaches have improved spatial reasoning along three paradigms. \textbf{Vision-centric methods} post-train MFMs on large-scale 3D-grounded samples, encouraging models to infer depth, size, location, or spatial relations directly from limited visual observations~\cite{spatialrgpt,spatialvlm,ma2025spatialreasoner,wu2025spatial}. While effective under in-distribution settings, recent studies~\cite{qi2025beyond,huang2025surprise3d,wu2025indoor} indicate that their gains are often coupled to the statistics of the training distribution and may degrade under shifts in scene layout. \textbf{Geometry-centric methods} inject explicit spatial signals, such as depth maps, point clouds, or learned 3D representations, to compensate for the weak geometric grounding of 2D visual inputs~\cite{daxberger2025mm,chen2025sd,ning2025enhancing,wang2025n3d}. Although useful, such signals introduce nontrivial cross-modal alignment challenges~\cite{zhang2025point} and often require geometry-rich paired data and costly training for specialized architectures. More recently, \textbf{tool-augmented spatial reasoning} enables models to call tools or task-specific models (e.g., 3D reconstruction, novel-view rendering, depth estimation, or pose estimation) for active evidence acquisition~\cite{zhang2026think3d,luo2026pyspatial,chen2025spacetools}. However, these tools typically return pixel-space observations or low-level geometric measurements, leaving the model to assemble high-level spatial structure through a long, error-prone reasoning chain. As a result, noisy or incomplete tool evidence can be absorbed rather than challenged, leading to incorrect reasoning results.

Indeed, biological intelligence (BI) offers a natural blueprint for overcoming the spatial reasoning bottleneck. Decades of cognitive science research suggest that BI does not passively match each incoming egocentric observation in isolation; instead, it compresses local perceptual experience into stable allocentric cognitive maps~\cite{burgess2006spatial}, supporting mental simulation~\cite{o1978hippocampus,schiller2015memory,bellmund2018navigating}. This suggests a sharper hypothesis: \textit{\textbf{robust spatial reasoning is constrained not merely by coarse 2D perception or limited 3D supervision, but by the absence of an allocentric spatial cognition framework that can be actively invoked, precisely queried, and cross-validated by foundation models.}} Building on this view, we propose \textbf{AlloSpatial}, an agentic framework that enables reasoning over structured allocentric spatial representations rather than relying solely on egocentric observations.

As illustrated in Fig.~\ref{fig:teaser}(A), at the core of AlloSpatial is \textbf{World2Mind}, a plug-and-play cognitive mapping sandbox that transforms egocentric videos or images into queryable allocentric spatial priors. Given semantic categories specified by the agent, World2Mind integrates a robust semantic-geometry alignment pipeline to construct a sparse semantic point cloud. It then distills this point cloud into two complementary cognitive maps: a \textbf{Landmark Cognitive Map} for object-centric topological reasoning and a \textbf{Route Cognitive Map} for traversability, passability, and trajectory reasoning. Its central representation is the proposed \textbf{Allocentric-Spatial Tree (AST)}, a directed acyclic graph whose nodes correspond to stable environmental landmarks and whose attributes encode centroid, footprint, principal axes, orientation, height range, and hierarchical containment. Unlike grid maps that discard object identity~\cite{yang2025thinking,wang2025mindcube,ruan2025reactive} or abstract semantic graphs that omit metric geometry~\cite{gu2024conceptgraphs,rana2023sayplan}, AST compresses noisy 3D reconstruction into compact and structured spatial memory.

However, structured spatial priors alone do not guarantee reliable reasoning, especially when reconstruction drift and perception errors often occur. AlloSpatial therefore introduces a three-stage \textbf{Spatial Reasoning Harness} to regulate how models invoke tools, collect evidence, and arbitrate across modalities. The harness first determines whether a question truly requires cognitive mapping, then decouples evidence collection from raw visual inputs, AST-structured text, and optional top-down visualization maps produced by World2Mind, and finally performs geometry-semantics interleaved reasoning to identify conflicts and cross-validate the final answer. To internalize the reasoning harness, we train an AlloSpatial agent instantiated from Qwen3-VL (see Fig.~\ref{fig:teaser}(B)). We first use World2Mind within the harness prompt to distill high-quality trajectories from frontier proprietary models, and then apply supervised cold-start fine-tuning to bootstrap tool invocation and reasoning structure. The agent is subsequently optimized with Group Sequence Policy Optimization (GSPO)~\cite{zheng2025group}. Since RL over long tool-use trajectories is vulnerable to reward hacking and redundant tool calls, we introduce a \textbf{Harness-Gated Trajectory Reward (HGTR)} that evaluates the trajectory as a whole rather than isolated actions. HGTR jointly accounts for answer correctness, harness compliance, tool-use validity, and response efficiency, while enforcing two key gates: answer accuracy is credited only when the trajectory follows the required reasoning structure, and tool-use rewards are granted only when valid tool calls contribute to a correct answer. This training strategy stabilizes optimization and enables the agent to acquire efficient and deliberate allocentric spatial reasoning behaviors.

We evaluate AlloSpatial on VSI-Bench~\cite{yang2025thinking} and MindCube~\cite{wang2025mindcube}, covering various spatial reasoning tasks across egocentric videos and sparse multi-view images. As a training-free plug-in, World2Mind combined with the Spatial Reasoning Harness consistently improves frontier commercial models, including GPT-5.2, Claude-4.6, and Gemini-3, with overall gains of approximately 5\%-18\%. Under a  ``blind'' setting in which visual inputs are removed, ASTs alone achieve strong 3D reasoning performance, suggesting that high-quality allocentric priors can elicit spatial mental simulation even without direct visual evidence. After cold-start RL, AlloSpatial agents instantiated from Qwen3-VL surpass larger frontier models and competitive methods across multiple tasks. In summary, our study indicates that coupling structured allocentric representations with an agentic reasoning harness offers a promising route toward overcoming the spatial cognition bottleneck of foundation models.


\section{Related Work: Spatial Reasoning in Foundation Models}

Recent benchmarks have exposed spatial reasoning as a persistent weakness of MFMs. VSI-Bench~\cite{yang2025thinking} and VSI-Super~\cite{yang2025cambrian} evaluate egocentric video-based spatial intelligence. MindCube~\cite{wang2025mindcube} probes sparse multi-view cognitive mapping, while other benchmarks~\cite{lin2025mmsi,ramakrishnan2024does} extend evaluation to broader text-based and multimodal scenarios. Together, these studies suggest that the bottleneck is not merely visual recognition, but the absence of reasoning-ready spatial mental representations.

\textbf{Vision-centric learning.}
Vision-centric methods improve spatial reasoning by post-training MFMs with large-scale 3D-grounded supervision. SpatialVLM~\cite{spatialvlm}, Cambrian-S~\cite{yang2025cambrian}, and related works~\cite{ouyang2025spatial,wang2025mindcube,deng2025internspatial} construct spatial VQA data from real-world images, videos, or 3D annotations to supervise spatial relation understanding. SpatialReasoner~\cite{ma2025spatialreasoner} and Spatial-MLLM~\cite{wu2025spatial} further introduce intermediate 3D representations or explicit reasoning traces to structure spatial inference. These methods show that spatially grounded supervision can improve benchmark performance. Still, their gains may remain coupled to training-distribution statistics and degrade under shifts in scene layout, viewpoint, or object composition. Recent analyses further suggest that MFMs can exploit semantic shortcuts rather than acquire robust spatial cognition~\cite{qi2025beyond,huang2025surprise3d,wu2025indoor}.

\textbf{Geometry-centric learning.}
Geometry-centric methods compensate for the weak geometric grounding of 2D inputs by injecting explicit spatial signals. SpatialBot~\cite{cai2025spatialbot} uses RGB-D inputs and depth-centered QA data to improve metric spatial understanding. MM-Spatial~\cite{daxberger2025mm} and SD-VLM~\cite{chen2025sd} study how depth maps, multi-view observations, and depth-encoded visual features affect 3D reasoning, while N3D-VLM~\cite{wang2025n3d} integrates native 3D grounding into a unified MFM. Other works explore point-cloud-enhanced LLMs or reasoning-based segmentation for localization and scene understanding~\cite{ning2025enhancing,zhang2025point}. Although explicit geometry provides useful spatial cues, directly conditioning MFMs on 3D modalities introduces cross-modal alignment challenges and often requires geometry-rich paired data, specialized architectures, and costly training.

\textbf{Tool-augmented learning.}
Tool-augmented methods equip MFMs with external modules for active evidence acquisition. Think3D~\cite{zhang2026think3d} and pySpatial~\cite{luo2026pyspatial} invoke 3D reconstruction, camera pose estimation, or novel-view rendering for interactive spatial exploration. SpaceTools~\cite{chen2025spacetools} studies how models coordinate depth, segmentation, and pose tools through reinforcement learning, while SpatialDreamer~\cite{cao2025spatialdreamer} uses tool hints or visual imagination as intermediate evidence. These approaches mark an important shift from passive perception to active tool use. However, most remain observation-centric: tools typically return rendered views, masks, depth maps, or low-level measurements, leaving the model to assemble high-level spatial structure through long and error-prone reasoning chains. In contrast, AlloSpatial converts noisy 3D reconstructions into \textbf{compact allocentric representations} and couples them with a Spatial Reasoning Harness for tool invocation, modality-decoupled evidence collection, and geometry--semantics arbitration. This enables models to reason over structured allocentric spatial memory rather than isolated egocentric observations or low-level geometric cues.

\section{Methodology}


\subsection{Problem Formulation}\label{subsec:formulation}

Given an egocentric observation sequence $\mathcal{I}=\{I_t\}_{t=1}^{T}$ and a spatial question $q$, our goal is to produce a final answer $\hat{a}$ together with a verifiable spatial reasoning process. We formulate AlloSpatial as a harness-guided tool-using agent $(\pi_\theta, \mathcal{W},\mathcal{H})$,
where $\pi_\theta$ denotes the foundation-model policy, $\mathcal{W}$ is the proposed World2Mind cognitive mapping sandbox, and $\mathcal{H}$ is the Spatial Reasoning Harness. Given $(q,\mathcal{I})$, the agent generates a multi-turn trajectory:
\begin{equation}
    \tau = \big(\mathbf{h}_0,\, u_0,\, o_0,\, \mathbf{h}_1,\, u_1,\, o_1,\, \dots,\, \mathbf{h}_K,\, \hat{a}\big),
\end{equation}
where $\mathbf{h}_k$ is the dialogue history at round $k$, and $u_k\sim\pi_\theta(\cdot\mid\mathbf{h}_k)$ is either a natural-language reasoning step or a structured tool call. If $u_k$ is a valid tool call, World2Mind returns an allocentric spatial observation $o_k=\mathcal{W}(u_k;\mathcal{I})$; otherwise, $o_k=\varnothing$. The final answer $\hat{a}$ is emitted within a predefined answer tag for reliable parsing and evaluation.

The harness $\mathcal{H}$ constrains the trajectory by specifying the required evidence channels, their ordering, and the cross-modal arbitration steps before final prediction. It therefore turns external allocentric priors into executable and inspectable reasoning evidence, rather than treating tool outputs as unverified context. For a training-free plug-in setting, $\mathcal{H}$ is instantiated via prompting. While in the training process, $\mathcal{H}$ defines the trajectory structure optimized during RL, as described in Sec.~\ref{subsec:train}. Thus, $\mathcal{H}$ serves both as an inference-time protocol and as a training-time inductive bias.

\subsection{World2Mind: Allocentric Cognitive Mapping Sandbox}\label{subsec:world2mind}

Rather than reconstructing an exhaustive scene model, World2Mind exposes a query-conditioned cognitive mapping interface. Given an egocentric observation sequence $\mathcal{I}$, an open-vocabulary category set $\mathcal{C}$, a requested knowledge type $k\in\{\textit{landmark},\textit{route},\textit{both}\}$, a footprint format $f\in\{\textit{rectangle},\textit{ellipse}\}$, and a scene type $s\in\{\textit{indoor},\textit{outdoor}\}$, World2Mind returns $\mathcal{M}=\mathcal{W}(\mathcal{I};\,\mathcal{C},k,f,s)$,
where $\mathcal{M}$ denotes the structured allocentric spatial knowledge.
\subsubsection{Geometry-Semantic Alignment Pipeline}\label{subsec:gs-pipeline}

We first align geometry and semantics across the egocentric sequence. For each frame $I_t\in\mathbb{R}^{H\times W}$, we estimates a dense depth map $D_t\in\mathbb{R}^{H\times W}$ and camera pose $T_t\in SE(3)$ using monocular geometry models~\cite{lin2025depth,wang2025vggt}, and extracts open-vocabulary semantic masks $\{M_t^c\}_{c\in\mathcal{C}}$ with SAM~3~\cite{carion2025sam}. To suppress unreliable geometry near object boundaries, textureless regions, and poorly reconstructed views, we apply a two-level confidence filter to the predicted confidence map $\Gamma_t\in[0,1]^{H\times W}$:
\begin{equation}
V_t(u,v)=
\mathbf{1}\!\left[\Gamma_t(u,v)>\tau_{\mathrm{px}}\right]
\cdot
\mathbf{1}\!\left[\bar{\Gamma}_t>\tau_{\mathrm{frm}}\right],
\quad
\bar{\Gamma}_t=\frac{1}{HW}\sum_{u,v}\Gamma_t(u,v),
\end{equation}
where $(u,v)$ indexes an image pixel, $V_t(u,v)\in\{0,1\}$ is the validity mask, and $\tau_{\mathrm{px}}$ and $\tau_{\mathrm{frm}}$ are the pixel- and frame-level confidence thresholds. 

Pixels satisfying both constraints are back-projected into the world coordinate system, where $p_t(u,v)$ denotes the 3D point obtained from $D_t(u,v)$ and $T_t$. Aggregating valid points, semantic labels, and colors across frames yields a global semantic point cloud $\mathcal{P}=\{(p_i,s_i,\mathrm{rgb}_i)\}_{i=1}^{N}$, where $p_i\in\mathbb{R}^3$, $s_i\in\mathcal{C}$, and $N$ is the number of retained points. To remove sparse outliers caused by boundary leakage and multi-view misalignment, we compute a $K$-nearest-neighbor density score $\rho_i=\frac{1}{K}\sum_{j\in\mathcal{N}_K(i)}\|p_i-p_j\|_2^{-1}$ for each point, where $\mathcal{N}_K(i)$ denotes the $K$ nearest neighbors of $p_i$, and discard points below a category-specific density percentile. The resulting point cloud is sparse but geometrically reliable, serving as the substrate for allocentric map construction.

\subsubsection{Allocentric Mapping}\label{subsec:cogmap}

World2Mind compresses the filtered $\mathcal{P}$ into complementary allocentric representations.

\textbf{Landmark mapping with Allocentric-Spatial Tree.}
For each queried category $c\in\mathcal{C}$, World2Mind applies adaptive DBSCAN to separate object instances. Each instance cluster is represented as a node in the \textbf{Allocentric-Spatial Tree (AST)} $\mathcal{T}=(\mathcal{V},\mathcal{E})$, where $\mathcal{V}$ denotes landmark nodes and $\mathcal{E}$ encodes hierarchical containment or support relations. Unlike abstract scene graphs that omit metric geometry~\cite{gu2024conceptgraphs,rana2023sayplan} or grid maps that discard object identity~\cite{yang2025thinking,wang2025mindcube}, AST preserves both object semantics and explicit spatial structure. For each node $v\in\mathcal{V}$, World2Mind projects its supporting points onto the ground plane and fits a compact footprint:
\begin{equation}
\phi(v)=
\big(
c_x,c_z;\;
a,b,\theta;\;
h_{\min},h_{\max};\;
A,n
\big),
\end{equation}
where $(c_x,c_z)$ is the top-down centroid, $(a,b)$ are the semi-axes of the fitted ellipse, $\theta$ is its orientation, $(h_{\min},h_{\max})$ denotes the vertical extent, $A$ is the footprint area, and $n$ is the number of supporting points, used as a confidence proxy. Elliptical footprints are used by default for stable coarse occupancy, while an axis-aligned rectangular format is returned when nearest-boundary computation is required. The resulting AST is serialized into \texttt{YAML} and included in the structured spatial map $\mathcal{M}$. This representation turns noisy 3D reconstruction into compact, model-readable spatial memory: foundation models can directly parse object centers, sizes, orientations, and containment relations without specialized 3D adapters, while the coarse footprint parameterization remains robust to local reconstruction noise and reflects the compressed nature of cognitive maps~\cite{burgess2006spatial,bellmund2018navigating}.

\textbf{Route mapping.}
For traversability and path-related queries, World2Mind voxelizes points associated with traversable categories, such as floors, and projects them onto a top-down grid. Each grid cell is labeled as traversable, occupied, or unknown. The camera trajectory $\{T_t\}_{t=1}^{T}$, where $T_t$ denotes the camera pose of frame $I_t$, is projected into the same coordinate system to encode observed motion history and support route-level reasoning.

\textbf{Optional visual rendering.}
Beyond structured text, World2Mind can return allocentric visualizations as auxiliary global observations, including top-down AST layouts, route maps, and semantic segmentation maps. These renderings provide a compact visual summary of the global scene layout and serve as additional evidence during geometry--semantics arbitration.

\subsection{Spatial Reasoning Harness}\label{subsec:harness}

Structured allocentric priors are useful but not self-verifying. Directly conditioning on World2Mind output $\mathcal{M}$ can induce two failure modes: \textit{1) modality lock-in}, where the model prematurely commits to either visual appearance or AST text and ignores conflicting evidence~\cite{wangself,madaan2023self}; and \textit{2) over-trust in tool results}, where incomplete or noisy reconstructions are treated as ground truth. To mitigate these failures, we design the \textbf{Spatial Reasoning Harness} $\mathcal{H}$ as a cyclic protocol:
\begin{equation}
\mathcal{H}:\quad
\textsc{Judge} \rightarrow
\textsc{Collect} \rightarrow
\textsc{Arbitrate}
\rightarrow
\{\textsc{Refine}, \textsc{Answer}\}.
\end{equation}
At each cycle, the agent decides whether additional allocentric evidence is needed, collects evidence through decoupled channels, and arbitrates geometry-semantics conflicts before either refining the query or committing to the final answer.

\textbf{Stage I: tool invocation judgment.}
At reasoning cycle $r$, the agent determines whether the current history $\mathbf{h}_r$ is sufficient or whether World2Mind should be queried. Following the reason-then-act principle of ReAct-style agents~\cite{yao2023react}, it first states a spatial hypothesis and the rationale for tool use:
\begin{equation}
(d_r,\eta_r)=\textsc{Judge}(q,\mathbf{h}_r),
\quad d_r\in\{\textsc{Call},\textsc{Skip}\},
\end{equation}
where $d_r$ is the tool-use decision and $\eta_r$ is the current spatial hypothesis. The agent calls World2Mind only when the question requires metric measurement, route topology, viewpoint transformation, or other allocentric spatial evidence, preserving an initial hypothesis that can later be verified or rejected.

\textbf{Stage II: modality-decoupled cue collection.}
When $d_r=\textsc{Call}$, the agent generates query parameters $\xi_r=(\mathcal{C}_r,k_r,f_r,s_r)$ and obtains a structured spatial map $\mathcal{M}_r=\mathcal{W}(\mathcal{I};\xi_r)$, where $\mathcal{C}_r$ is the queried category set, $k_r$ the requested knowledge type, $f_r$ the footprint format, and $s_r$ the scene type. Evidence is then collected separately from raw visual observations, AST-structured text, and optional top-down maps $\{e_r^{\mathrm{vis}},e_r^{\mathrm{ast}},e_r^{\mathrm{map}}\}$.
Here, $\mathcal{E}_r$ denotes the evidence set at cycle $r$, and $e_r^{\mathrm{vis}}$, $e_r^{\mathrm{ast}}$, and $e_r^{\mathrm{map}}$ denote visual, textual-allocentric, and rendered-map evidence, respectively. This decoupled collection prevents premature fusion and reduces reliance on a single evidence channel.

\textbf{Stage III: geometry-semantics interleaved arbitration.}
The agent then compares visual semantics with geometric evidence from AST and optional maps, identifies conflicts caused by reconstruction drift, missing instances, or false detections, etc., and then decides whether the evidence is sufficient:
\begin{equation}
(\Delta_r,\omega_r,y_r)=\textsc{Arbitrate}(\mathcal{E}_r),
\quad y_r\in\{\textsc{Refine},\textsc{Answer}\},
\end{equation}
where $\Delta_r$ denotes detected cross-modal conflicts, $\omega_r$ represents confidence assignments over evidence channels, and $y_r$ is the next action. If $y_r=\textsc{Refine}$, the agent updates $\xi_r$ and starts another cycle; if $y_r=\textsc{Answer}$, it emits the final response within the \texttt{<Answer>} tag. In this way, $\mathcal{H}$ treats external cognitive maps as falsifiable evidence for reasoning rather than as direct pseudo-labels, enabling iterative, verifiable, and trainable allocentric spatial reasoning.

\subsection{Internalizing the Spatial Reasoning Harness via Reinforcement Learning}
\label{subsec:train}

While frontier proprietary models can often follow the Spatial Reasoning Harness $\mathcal{H}$ through prompting, open-weight models~\cite{bai2025qwen3,wang2025internvl3} do not naturally exhibit stable multi-turn tool-use behavior. In preliminary experiments, they frequently produce malformed tool calls, skip cross-modal arbitration, or degenerate into repetitive interaction. We therefore internalize $\mathcal{H}$ into the open-weight policy $\pi_\theta$ with supervised cold start followed by RL with live World2Mind execution.

\begin{table*}[t]
\centering
\caption{Results of AlloSpatial under the training-free setting on the VSI-Bench~\cite{yang2025thinking} and MindCube~\cite{wang2025mindcube} benchmarks (tiny split). For VSI-Bench, we uniformly use 32 input frames.}
\label{table:infer_only_results}
\renewcommand{\arraystretch}{1.0}
\setlength{\tabcolsep}{2.2pt}
\scriptsize
\begin{adjustbox}{width=0.95\textwidth}
\begin{tabular}{@{}l ccccccccc cccc@{}}
\toprule
\multirow{2}{*}{\textbf{Models}}
& \multicolumn{9}{c}{\textbf{VSI-Bench}}
& \multicolumn{4}{c}{\textbf{MindCube}} \\
\cmidrule(lr){2-10} \cmidrule(l){11-14}
& \rotatebox{60}{\textbf{Overall}}
& \rotatebox{60}{\textbf{Obj. Count}}
& \rotatebox{60}{\textbf{Abs. Dist.}}
& \rotatebox{60}{\textbf{Obj. Size}}
& \rotatebox{60}{\textbf{Room Size}}
& \rotatebox{60}{\textbf{Rel. Dist.}}
& \rotatebox{60}{\textbf{Rel. Dir.}}
& \rotatebox{60}{\textbf{Route Plan}}
& \rotatebox{60}{\textbf{Appr. Order}}
& \rotatebox{60}{\textbf{Overall}}
& \rotatebox{60}{\textbf{Around}}
& \rotatebox{60}{\textbf{Among}}
& \rotatebox{60}{\textbf{Rotation}} \\
\midrule
\rowcolor{propriHdr!75}
\multicolumn{14}{c}{\textbf{\textit{Proprietary Foundation Models w/o. AlloSpatial}}} \\

\modelicon{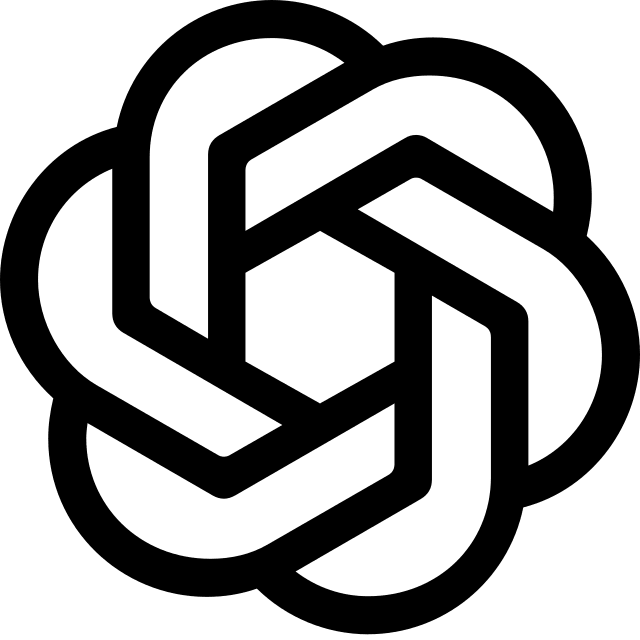}~GPT-5.2
& 46.7 & 52.5 & 34.9 & 67.5 & 50.6 & 42.0 & 40.7 & 34.7 & 51.0
& 49.9 & 62.4 & 45.2 & 48.5 \\

\modelicon{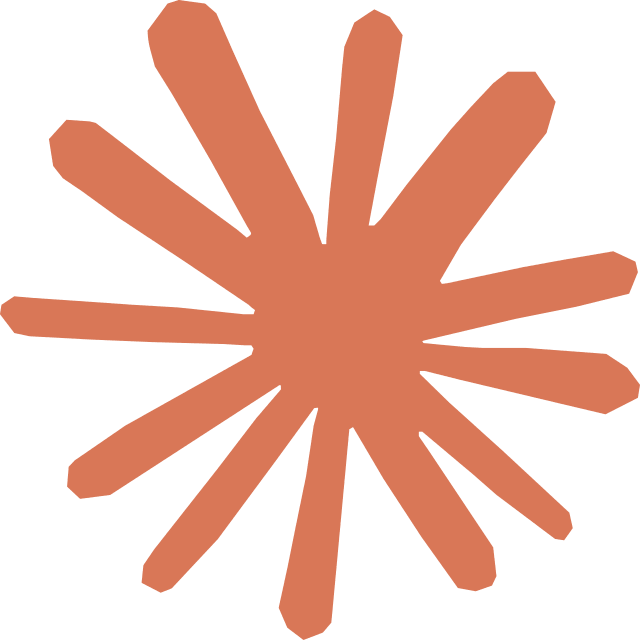}~Claude-4.6-Opus
& 38.4 & 46.9 & 18.5 & 62.1 & 26.8 & 40.0 & 47.2 & 34.7 & 30.6
& 48.5 & 58.8 & 50.7 & 29.0 \\

\modelicon{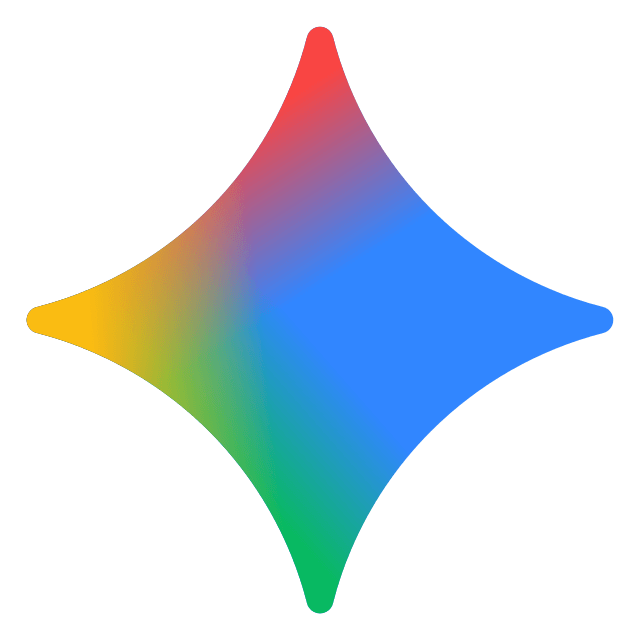}~Gemini-3-Pro
& 55.2 & 47.8 & 32.1 & \textbf{71.3} & 55.0 & 54.0 & 44.8 & 57.1 & 79.6
& 75.1 & 77.2 & 68.2 & 93.0 \\

\midrule
\rowcolor{opensrcHdr!75}
\multicolumn{14}{c}{\textbf{\textit{Proprietary Foundation Models w./ AlloSpatial}}} \\

\multirow{2}{*}{\modelicon{icons/gpt.png}~GPT-5.2}
& 54.0 & 47.4 & 33.4 & 63.3 & 52.4 & \textbf{64.0} & 41.1 & 51.0 & 79.6
& 54.6 & 60.4 & 47.7 & 68.0 \\
& {\scriptsize \textcolor{green!60!black}{($\uparrow$7.3)}}
& {\scriptsize \textcolor{red}{($\downarrow$5.1)}}
& {\scriptsize \textcolor{red}{($\downarrow$1.5)}}
& {\scriptsize \textcolor{red}{($\downarrow$4.2)}}
& {\scriptsize \textcolor{green!60!black}{($\uparrow$1.8)}}
& {\scriptsize \textcolor{green!60!black}{($\uparrow$22.0)}}
& {\scriptsize \textcolor{green!60!black}{($\uparrow$0.4)}}
& {\scriptsize \textcolor{green!60!black}{($\uparrow$16.3)}}
& {\scriptsize \textcolor{green!60!black}{($\uparrow$28.6)}}
& {\scriptsize \textcolor{green!60!black}{($\uparrow$4.7)}}
& {\scriptsize \textcolor{red}{($\downarrow$2.0)}}
& {\scriptsize \textcolor{green!60!black}{($\uparrow$2.5)}}
& {\scriptsize \textcolor{green!60!black}{($\uparrow$19.5)}} \\

\addlinespace[1pt]
\multirow{2}{*}{\modelicon{icons/claude.png}~Claude-4.6-Opus}
& 56.0 & \textbf{59.0} & 34.3 & 67.3 & 54.8 & \textbf{64.0} & 62.7 & \textbf{65.3} & 40.8
& 62.9 & 82.4 & 60.8 & 45.0 \\
& {\scriptsize \textcolor{green!60!black}{($\uparrow$17.7)}}
& {\scriptsize \textcolor{green!60!black}{($\uparrow$12.0)}}
& {\scriptsize \textcolor{green!60!black}{($\uparrow$15.8)}}
& {\scriptsize \textcolor{green!60!black}{($\uparrow$5.2)}}
& {\scriptsize \textcolor{green!60!black}{($\uparrow$28.0)}}
& {\scriptsize \textcolor{green!60!black}{($\uparrow$24.0)}}
& {\scriptsize \textcolor{green!60!black}{($\uparrow$15.6)}}
& {\scriptsize \textcolor{green!60!black}{($\uparrow$30.6)}}
& {\scriptsize \textcolor{green!60!black}{($\uparrow$10.2)}}
& {\scriptsize \textcolor{green!60!black}{($\uparrow$14.4)}}
& {\scriptsize \textcolor{green!60!black}{($\uparrow$23.6)}}
& {\scriptsize \textcolor{green!60!black}{($\uparrow$10.1)}}
& {\scriptsize \textcolor{green!60!black}{($\uparrow$16.0)}} \\

\addlinespace[1pt]
\multirow{2}{*}{\modelicon{icons/gemini.png}~Gemini-3-Pro}
& \textbf{61.0} & 51.8 & \textbf{36.8} & 57.7 & \textbf{62.6} & 62.0 & \textbf{67.7} & \textbf{65.3} & \textbf{83.7}
& \textbf{81.6} & \textbf{86.0} & \textbf{75.8} & \textbf{93.5} \\
& {\scriptsize \textcolor{green!60!black}{($\uparrow$5.8)}}
& {\scriptsize \textcolor{green!60!black}{($\uparrow$4.1)}}
& {\scriptsize \textcolor{green!60!black}{($\uparrow$4.7)}}
& {\scriptsize \textcolor{red}{($\downarrow$13.5)}}
& {\scriptsize \textcolor{green!60!black}{($\uparrow$7.6)}}
& {\scriptsize \textcolor{green!60!black}{($\uparrow$8.0)}}
& {\scriptsize \textcolor{green!60!black}{($\uparrow$22.9)}}
& {\scriptsize \textcolor{green!60!black}{($\uparrow$8.2)}}
& {\scriptsize \textcolor{green!60!black}{($\uparrow$4.1)}}
& {\scriptsize \textcolor{green!60!black}{($\uparrow$6.5)}}
& {\scriptsize \textcolor{green!60!black}{($\uparrow$8.8)}}
& {\scriptsize \textcolor{green!60!black}{($\uparrow$7.6)}}
& {\scriptsize \textcolor{green!60!black}{($\uparrow$0.5)}} \\

\bottomrule
\end{tabular}
\end{adjustbox}
\vspace{-0.3cm}
\end{table*}

\textbf{Supervised cold start.}
We first use World2Mind and the harness protocol to distill trajectories from proprietary models. We retain trajectories that are answer-correct, structurally valid, and contain non-trivial cross-modal arbitration. Supervised fine-tuning on these traces bootstraps tool-call syntax, AST and route-map parsing, harness stage ordering, and answer-tag formatting, thereby preparing the policy for subsequent optimization with Group Sequence Policy Optimization (GSPO)~\cite{zheng2025group}.

\textbf{GSPO with Harness-Gated Trajectory Reward.}
To stabilize reinforcement learning over long tool-use trajectories, we introduce a \textbf{Harness-Gated Trajectory Reward (HGTR)} that scores each complete trajectory $\tau$ rather than individual tokens. HGTR combines answer correctness, structural compliance, tool-use effectiveness, and response efficiency:
\begin{equation}
R(\tau)
=
w_{\mathrm{acc}} \widetilde R_{\mathrm{acc}}(\tau)
+
w_{\mathrm{str}} R_{\mathrm{str}}(\tau)
+
w_{\mathrm{tool}} R_{\mathrm{tool}}(\tau)
+
w_{\mathrm{len}} R_{\mathrm{len}}(\tau),
\end{equation}
where $w_{\mathrm{acc}},w_{\mathrm{str}},w_{\mathrm{tool}},w_{\mathrm{len}}$ are reward weights. $R_{\mathrm{str}}(\tau)$ measures harness compliance, including valid tool-call tags, complete stage ordering, explicit arbitration, and the final answer tag. $R_{\mathrm{acc}}(\tau)$ measures answer quality, using exact match for multiple-choice questions and mean relative accuracy for numerical questions following~\cite{yang2025thinking}. To prevent malformed trajectories from receiving reward through accidental correct guesses, HGTR applies a \textbf{structure-gated accuracy}:
\begin{equation}
\widetilde R_{\mathrm{acc}}(\tau)
=
R_{\mathrm{acc}}(\tau)
\cdot
\mathbf{1}\!\left[
R_{\mathrm{str}}(\tau)\ge \tau_{\mathrm{s}}
\right],
\end{equation}
where $\tau_{\mathrm{s}}$ is the threshold and $\mathbf{1}[\cdot]$ is the indicator function. Thus, correctness is credited only when the trajectory satisfies the harness format. HGTR further uses a \textbf{correctness-tied tool-use reward}:
\begin{equation}
R_{\mathrm{tool}}(\tau)
=
\alpha\,\widetilde R_{\mathrm{acc}}(\tau)
\mathbf{1}\!\left[\mathrm{ValidCall}(\tau)\right]
-
\gamma\,
\max\!\left(0,\, n_{\mathcal{W}}(\tau)-n^\star\right),
\end{equation}
where $\mathrm{ValidCall}(\tau)$ indicates whether World2Mind is invoked with valid syntax and task-relevant arguments, $n_{\mathcal{W}}(\tau)$ is the number of calls, $n^\star$ is a soft call budget, and $\alpha,\gamma$ control the tool reward and overuse penalty. This reward grants tool-use credit only when a valid invocation contributes to a structurally valid and correct answer, while discouraging redundant reconstruction once sufficient allocentric evidence is available. Finally, $R_{\mathrm{len}}(\tau)$ penalizes excessive model-generated tokens while masking out tool-returned ASTs, route maps, and visualization metadata. Overall, HGTR guides the agent toward trajectories that are answer-correct, harness-compliant, tool-efficient, and verifiable.

\begin{table*}[t]
\centering
\caption{Evaluation results on the VSI-Bench benchmark (tiny split).
Following the setting of Think3D~\cite{zhang2026think3d}, all models are evaluated with \textbf{7} input frames to assess spatial reasoning under limited egocentric observations. \dag: Results are directly taken from the original paper. Green shades denote the \topone, \toptwo, and \topthree performance within each metric.}
\label{table:vsibench_tiny_full}
\renewcommand{\arraystretch}{1.0}
\setlength{\tabcolsep}{2.0pt}
\scriptsize
\begin{adjustbox}{width=\textwidth}
\begin{tabular}{@{}c l c ccccc ccccc@{}}
\toprule
\multirow{3}{*}{\textbf{Rank}} 
& \multirow{3}{*}{\textbf{Models}} 
& \multirow{3}{*}{\textbf{Overall}} 
& \multicolumn{5}{c}{\textbf{Numerical Answer (MRA)}} 
& \multicolumn{5}{c}{\textbf{Multiple-Choice Answer (Pass@1 Accuracy)}} \\
\cmidrule(lr){4-8} \cmidrule(l){9-13}
& & 
& \textbf{Avg.} 
& \textbf{Obj. Count} 
& \textbf{Abs. Dist.} 
& \textbf{Obj. Size} 
& \textbf{Room Size} 
& \textbf{Avg.} 
& \textbf{Rel. Dist.} 
& \textbf{Rel. Dir.} 
& \textbf{Route Plan} 
& \textbf{Appr. Order} \\
\midrule

\rowcolor{propriHdr!75}
\multicolumn{13}{c}{\textbf{\textit{Proprietary Foundation Models}}} \\

11 & \modelicon{icons/gpt.png}~GPT-5.2 
& 37.7 & 36.4 & 28.6 & 24.3 & 52.3 & 40.6 
& 39.0 & 42.0 & 28.4 & \thirdcell{42.9} & 42.9 \\

8 & \modelicon{icons/gemini.png}~Gemini-2.5-Pro
& 46.2 & 39.4 & 34.7 & 14.3 & 56.3 & 52.4
& \thirdcell{53.0} & \thirdcell{52.0} & 45.7 & 38.8 & \secondcell{75.5} \\

9 & \modelicon{icons/gemini.png}~Gemini-3-Pro 
& 45.2 & 32.2 & 29.4 & 12.6 & 38.3 & 48.4 
& \secondcell{58.3} & \bestcell{56.0} & \thirdcell{52.7} & \bestcell{49.0} & \secondcell{75.5} \\

\midrule
\rowcolor{opensrcHdr!75}
\multicolumn{13}{c}{\textbf{\textit{Open-source Foundation Models}}} \\

12 & \modelicon{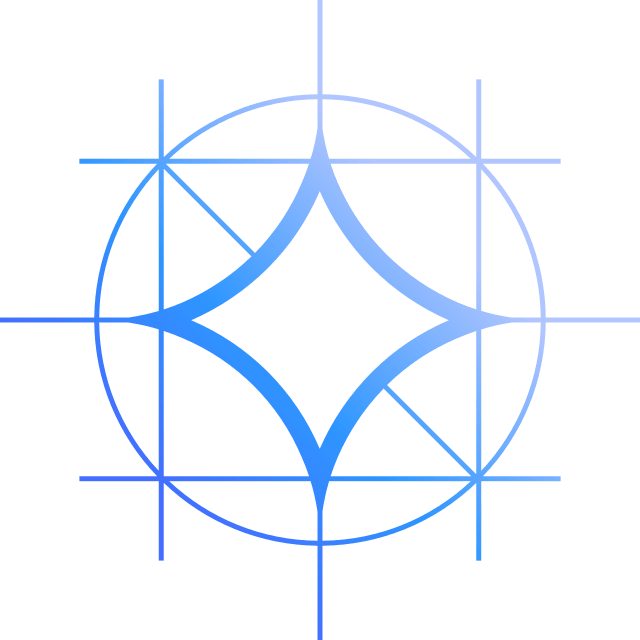}~Gemma-4-E4B 
& 33.5 & 30.1 & 19.6 & 24.5 & 33.1 & 43.2 
& 36.9 & 38.0 & 40.0 & 30.6 & 38.8 \\

6 & \modelicon{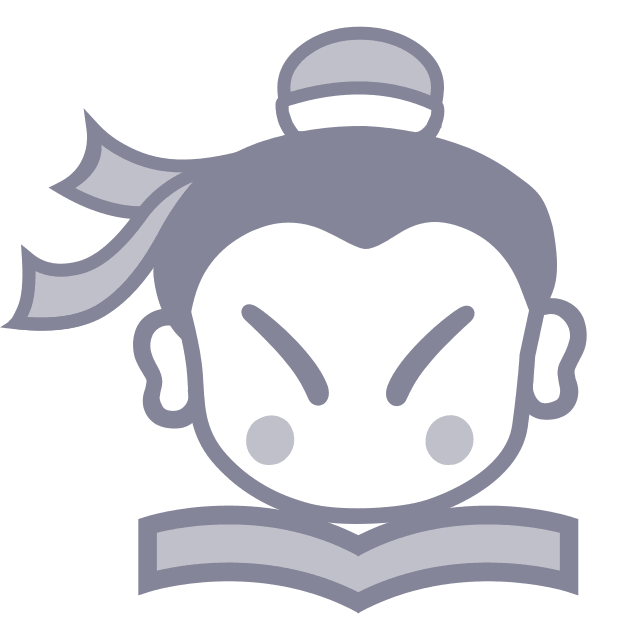}~InternVL3.5-4B 
& 48.8 & 50.9 & \bestcell{77.3} & 19.4 & 60.4 & 46.4 
& 46.8 & 38.0 & 47.1 & 40.8 & 61.2 \\

10 & \modelicon{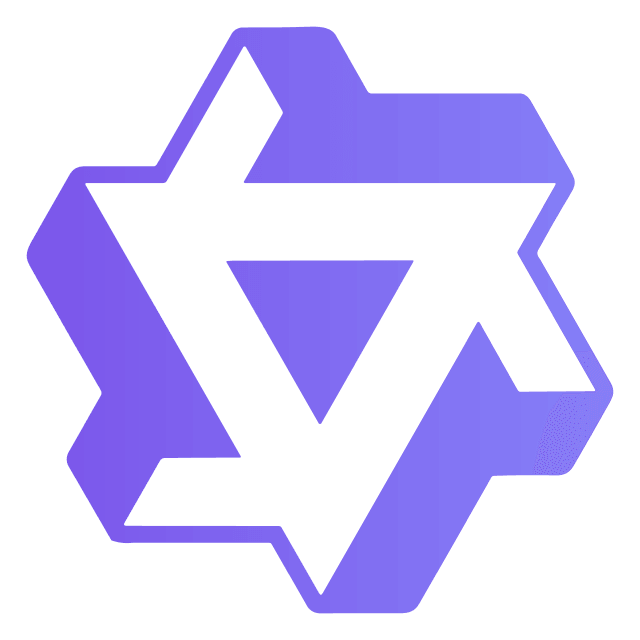}~Qwen3-VL-4B 
& 45.1 & 47.7 & 44.3 & 34.5 & 64.4 & 47.8 
& 42.4 & 36.0 & 37.8 & 24.5 & \thirdcell{71.4} \\

7 & \modelicon{icons/qwen.png}~Qwen3-VL-8B 
& 48.5 & 51.5 & 50.4 & \secondcell{40.2} & 67.5 & 48.0 
& 45.5 & 36.0 & 44.1 & 30.6 & \thirdcell{71.4} \\

3 & \modelicon{icons/qwen.png}~Qwen3-VL-32B 
& \thirdcell{53.1} & \secondcell{55.3} & 60.6 & \thirdcell{35.1} & \secondcell{71.5} & \secondcell{54.0} 
& 50.9 & 46.0 & 51.4 & 36.7 & 69.4 \\

\midrule
\rowcolor{gray!12}
\multicolumn{13}{c}{\textbf{\textit{Specialized Spatial Models}}} \\

5 & Spatial-MLLM-4B~\cite{wu2025spatial} 
& 49.1 & \thirdcell{53.0} & \secondcell{76.5} & 27.0 & 57.9 & 50.6 
& 45.2 & 44.0 & 40.9 & 38.8 & 57.1 \\

4 & Cambrian-S-3B~\cite{yang2025cambrian} 
& 49.5 & 51.7 & \thirdcell{67.1} & 22.6 & \bestcell{73.8} & 43.2 
& 47.3 & \secondcell{54.0} & 37.1 & 26.5 & \thirdcell{71.4} \\

-- & Think3D-4B\dag~\cite{zhang2026think3d} 
& $-$ & $-$ & $-$ & $-$ & $-$ & $-$ 
& 45.4 & 44.7 & 39.0 & 36.7 & 61.2 \\

2 & \modelicon{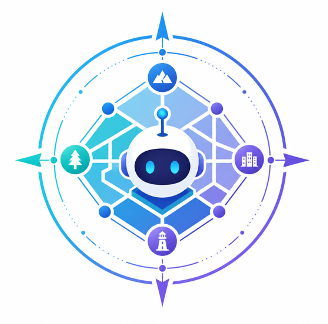}~AlloSpatial-4B (Ours) 
& \secondcell{53.5} & \bestcell{56.3} & 54.1 & \bestcell{42.3} & \thirdcell{68.8} & \bestcell{60.0} 
& 50.8 & 50.0 & \secondcell{59.3} & 34.7 & 59.2 \\

1 & \modelicon{icons/allo.png}~AlloSpatial-8B (Ours) 
& \bestcell{54.2} & 47.8 & 46.1 & 26.0 & 65.6 & \thirdcell{53.4} 
& \bestcell{60.6} & \thirdcell{52.0} & \bestcell{64.1} & \secondcell{46.9} & \bestcell{79.6} \\

\bottomrule
\end{tabular}
\end{adjustbox}
\vspace{-0.5cm}
\end{table*}

\section{Experiments}
\subsection{Experimental Setup}

\textbf{Datasets.}
We evaluate AlloSpatial on the official Tiny split of VSI-Bench~\cite{yang2025thinking} and MindCube~\cite{wang2025mindcube}, containing 392 and 1,050 questions, respectively. For training, we sample data from VSI-590K and the MindCube training set. The cold-start stage distills multi-turn tool-use trajectories from GPT-5.2 and Claude-4.6-Opus with World2Mind and the Spatial Reasoning Harness, followed by filtering for answer correctness, harness compliance, and valid tool invocation.

\textbf{Metrics.}
Following~\cite{yang2025thinking,wang2025mindcube}, we report pass@1 accuracy for multiple-choice questions and mean relative accuracy (MRA) for numerical questions. The overall score is the unweighted average over all task types within each benchmark. For the training-free setting on VSI-Bench, we uniformly sample up to 32 frames from each video. For the post-trained AlloSpatial agents and other baselines, we follow the limited-observation setting of Think3D~\cite{zhang2026think3d} and uniformly sample 7 frames, testing performance under sparse visual inputs. We further analyze the effect of frame number in Sec.~\ref{sec:ablation}.

\textbf{Baselines.}
We compare \textit{1) Proprietary models} including GPT-5.2, Claude-4.6-Opus, and Gemini-3-Pro; \textit{2) Open-source models} including Qwen3-VL~\cite{bai2025qwen3}, InternVL3.5~\cite{wang2025internvl3}, and Gemma-4~\cite{googledeepmind2026gemma4}; \textit{3) Specialized spatial models} including Spatial-MLLM~\cite{wu2025spatial}, Cambrian-S~\cite{yang2025cambrian}, and Think3D~\cite{zhang2026think3d}.

\textbf{Training details.}
We instantiate AlloSpatial from Qwen3-VL-4B-Instruct and Qwen3-VL-8B-Instruct~\cite{bai2025qwen3}. The 4B and 8B agents are trained for 600 and 400 RL steps, respectively, using 4.8K and 2.4K unique prompts, with 8 sampled rollouts per prompt. Detailed hyperparameters, World2Mind service configuration, and computational costs are provided in the Appendices~\ref{appendix:config}, \ref{appendix:World2Mind}, and \ref{appendix:costs}.

\begin{figure*}[t]
\centering

\begin{minipage}[t]{0.55\textwidth}
\centering
\captionof{table}{Evaluation on the MindCube (tiny split).}
\label{table:mindcube_tiny_full}
\vspace{4pt}
\renewcommand{\arraystretch}{1.0}
\setlength{\tabcolsep}{3.2pt}
\scriptsize
\begin{adjustbox}{width=\linewidth}
\begin{tabular}{@{}c l c ccc@{}}
\toprule
\textbf{Rank} 
& \textbf{Models} 
& \textbf{Overall} 
& \textbf{Around} 
& \textbf{Among} 
& \textbf{Rotation} \\
\midrule

2 & Gemini-2.5-Pro 
& \secondcell{57.9} & \secondcell{67.2} & \secondcell{43.8} & \bestcell{88.5} \\

5 & Gemma-4-E4B 
& 37.3 & 37.6 & \thirdcell{38.0} & 35.0 \\

6 & InternVL3.5-4B 
& 36.6 & 41.6 & 35.3 & 34.0 \\

8 & Qwen3-VL-4B 
& 28.3 & 40.4 & 20.8 & 35.5 \\

4 & Spatial-MLLM-4B~\cite{wu2025spatial} 
& 39.5 & \thirdcell{52.0} & 36.3 & 33.5 \\

7 & Cambrian-S-3B~\cite{yang2025cambrian} 
& 33.2 & 34.4 & 34.5 & 28.0 \\

3 & Think3D-4B\dag~\cite{zhang2026think3d} 
& \thirdcell{44.0} & 42.5 & 37.5 & \thirdcell{42.5} \\

1 & AlloSpatial-4B (Ours) 
& \bestcell{69.1} & \bestcell{82.0} & \bestcell{65.0} & \secondcell{65.5} \\

\bottomrule
\end{tabular}
\end{adjustbox}
\end{minipage}
\hfill
\begin{minipage}[t]{0.38\textwidth}
\vspace{-3pt}
\centering
\includegraphics[width=\linewidth]{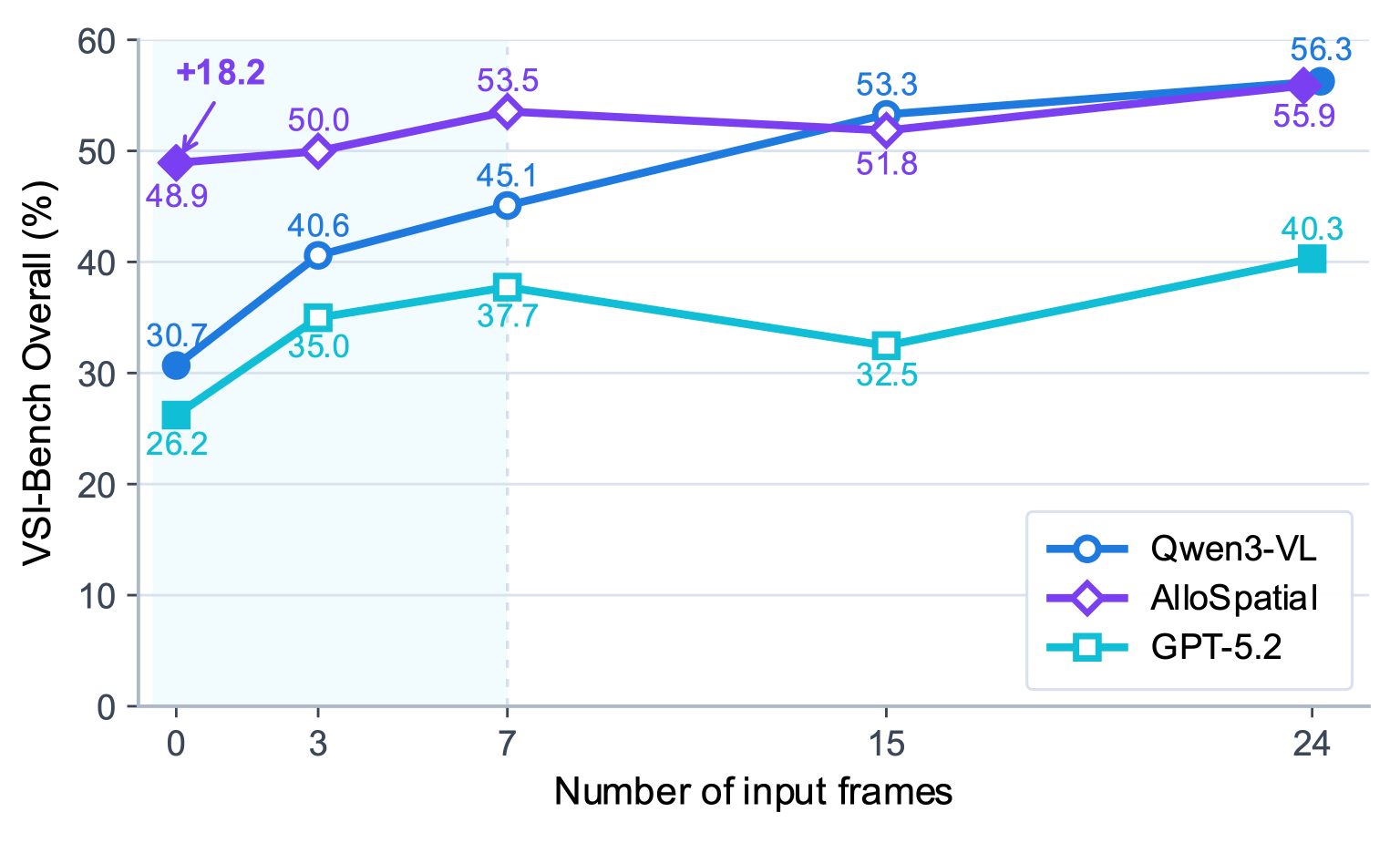}
\vspace{-0.5cm}
\captionof{figure}{VSI-Bench performance under different input-frame number.}
\label{fig:frame}
\end{minipage}
\vspace{-0.5cm}
\end{figure*}

\subsection{AlloSpatial Improves Proprietary Foundation Models for Spatial Reasoning}
\label{subsec:exp_frontier}

We first evaluate AlloSpatial as a \emph{training-free} plug-in for proprietary models by exposing World2Mind $\mathcal{W}$ and the Spatial Reasoning Harness $\mathcal{H}$ through prompting.

\textbf{Consistent improvement across benchmarks.} As shown in Tab.~\ref{table:infer_only_results}, AlloSpatial improves GPT-5.2, Claude-4.6-Opus, and Gemini-3-Pro on VSI-Bench by $+7.3$, $+17.7$, and $+5.8$ overall points, respectively, and on MindCube by $+4.7$, $+14.4$, and $+6.5$ points. The gains are concentrated on tasks that require allocentric structure, such as relative direction, route planning, and viewpoint-dependent rotation, while tasks solvable from local visual evidence show smaller or occasionally negative changes. This pattern suggests that World2Mind is most beneficial when direct egocentric perception is insufficient, and the model must reason over stable spatial relations beyond the observed view. A complete reasoning trace is shown in Fig.~\ref{fig:case}, and additional analysis for AlloSpatial's reasoning cases are illustrated in Appendix~\ref{appendix:case}.

\begin{wrapfigure}[12]{r}{0.40\textwidth}
    \centering
    \includegraphics[width=\linewidth]{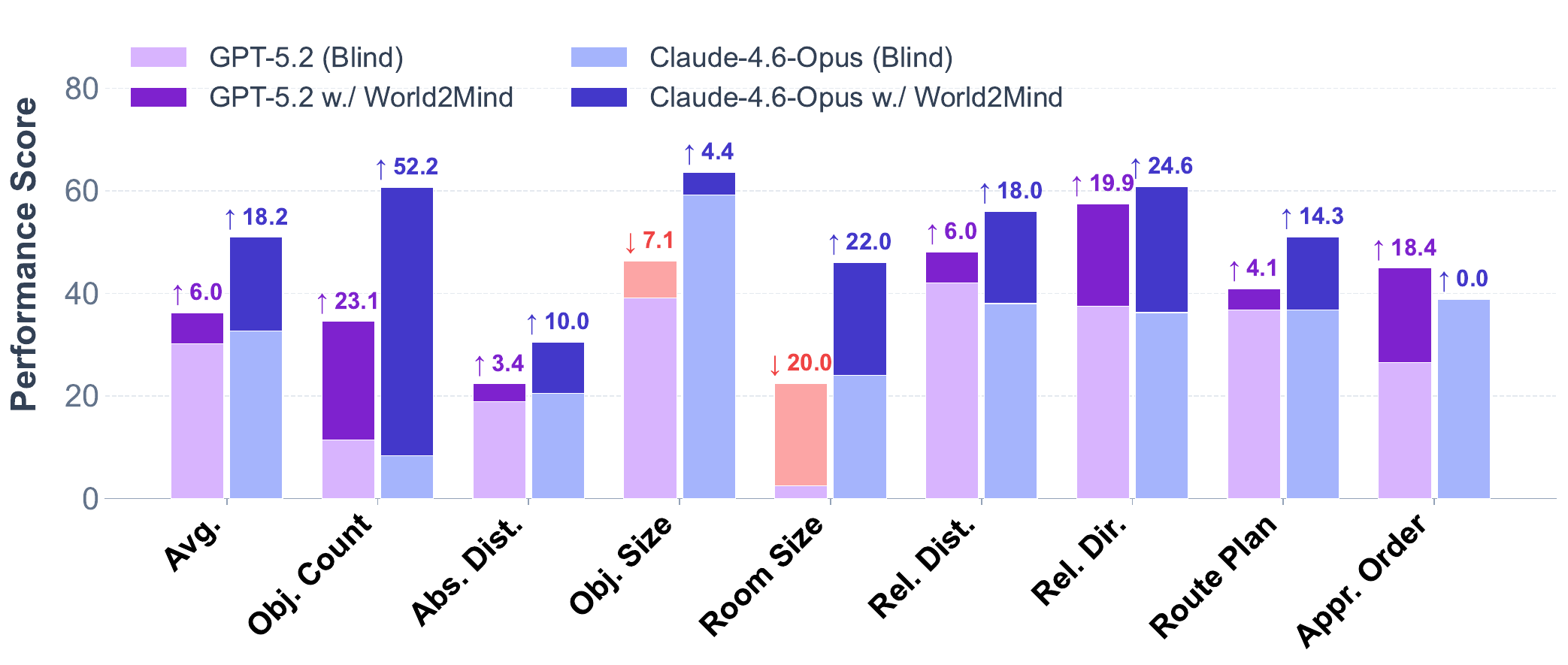}
    \caption{\textbf{Text-only spatial reasoning with allocentric priors.}
    VSI-Bench performance under the ``blind'' setting, where visual inputs are removed.}
    \label{fig:abla}
    \vspace{-0.8em}
\end{wrapfigure}

\textbf{ASTs provide effective allocentric reasoning evidence.} 
We further test whether the gain comes from structured spatial knowledge itself by evaluating VSI-Bench in a text-only ``blind'' setting~\cite{yang2025thinking}. Fig.~\ref{fig:abla} shows that AST text substantially improves blind spatial reasoning over text-only baselines, especially on object size and route planning. This supports the role of AST as a compact allocentric prior that enables models to reason over spatial structure without directly observing the scene. Together with the full-input results, the blind setting suggests that the critical signal is not merely additional visual context, but the structured spatial organization supplied by World2Mind.

\subsection{Trained AlloSpatial Agents Outperform General and Spatially Specialized Models} \label{subsec:exp_trained}

We next compare the trained AlloSpatial agents with proprietary models, open-source MFMs, and specialized spatial reasoning models. As shown in Tab.~\ref{table:vsibench_tiny_full}, AlloSpatial-8B achieves the best overall score on VSI-Bench, while AlloSpatial-4B ranks second. Both variants outperform proprietary models such as GPT-5.2, Gemini-2.5-Pro, and Gemini-3-Pro. Notably, AlloSpatial-4B and AlloSpatial-8B also surpass Qwen3-VL-32B, despite using substantially smaller backbones. AlloSpatial also compares favorably with spatially specialized models. On VSI-Bench, AlloSpatial-8B improves over Spatial-MLLM-4B and Cambrian-S-3B by $+5.1$ and $+4.7$ overall points, respectively, while AlloSpatial-4B also exceeds both baselines. This is notable because Spatial-MLLM and Cambrian-S rely on large-scale spatial grounding data, whereas AlloSpatial uses a much smaller set of unique training prompts and acquires spatial competence through structured tool interaction and trajectory-level optimization. These results suggest that allocentric spatial memory and verifiable reasoning can improve data efficiency compared with purely supervision-driven spatial learning.

Considering task-level performance, AlloSpatial shows greater improvements on relational and viewpoint-dependent tasks, such as relative direction, route planning, appearance order, and MindCube tasks. As illustrated in Tab.~\ref{table:mindcube_tiny_full}, AlloSpatial-4B reaches $69.1\%$ overall accuracy on MindCube, clearly outperforming Gemini-2.5-Pro, Think3D-4B, Spatial-MLLM-4B, and Cambrian-S-3B, with particularly large gains on \textsc{Around} and \textsc{Among}. In contrast, improvements on precise numerical estimation tasks, such as absolute distance and object counting, are less uniform. This pattern is consistent with the design of World2Mind: the AST and route maps provide robust coarse allocentric structure for relational reasoning, but the current reconstruction pipeline still limits fine-grained metric accuracy. Overall, \textbf{AlloSpatial agents are most effective when the task benefits from a stable allocentric organization rather than exact metric reconstruction.}

\begin{figure*}[t]
\centering

\begin{minipage}[t]{0.51\textwidth}
\vspace{0pt}
\centering
\includegraphics[width=\linewidth]{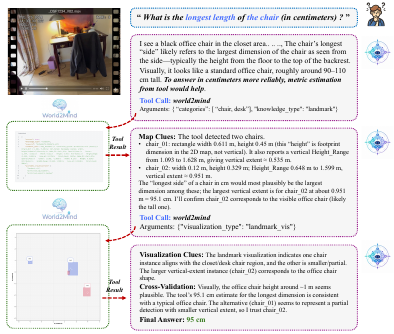}
\vspace{-0.5cm}
\captionof{figure}{Reasoning trace produced by AlloSpatial.}
\label{fig:case}
\end{minipage}
\hfill
\begin{minipage}[t]{0.45\textwidth}
\vspace{0pt}
\centering
\captionof{table}{\textbf{Ablation results of AlloSpatial-4B.}
We compare Qwen3-VL-4B variants using the same training data as QA pairs for SFT and RL, and report AlloSpatial-4B at Stage-1 (SFT cold start) and Stage-2 (RL).}
\label{tab:ablation}
\vspace{0.2cm}
\renewcommand{\arraystretch}{1.05}
\setlength{\tabcolsep}{3.2pt}
\scriptsize
\begin{adjustbox}{width=\linewidth}
\begin{tabular}{@{}l c c c@{}}
\toprule
\textbf{Variants} 
& \textbf{VSI-Bench} 
& \textbf{MindCube} 
& \textbf{Tokens} \\
\midrule

\rowcolor{propriHdr!55}
\multicolumn{4}{c}{\textbf{\textit{Qwen3-VL-4B}}} \\

\textit{Instruct}
& 45.1 & 28.3 & -- \\

\textit{Thinking}
& 45.5 {\tiny \textcolor{green!60!black}{($\uparrow$0.4)}}
& 36.1 {\tiny \textcolor{green!60!black}{($\uparrow$7.8)}}
& 1064 \\

\textit{w./ SFT on QAs}
& 43.1 {\tiny \textcolor{red}{($\downarrow$2.0)}}
& 53.9 {\tiny \textcolor{green!60!black}{($\uparrow$25.6)}}
& -- \\

\textit{w./ RL on QAs}
& 46.2 {\tiny \textcolor{green!60!black}{($\uparrow$1.1)}}
& 53.0 {\tiny \textcolor{green!60!black}{($\uparrow$24.7)}}
& -- \\

\midrule

\rowcolor{opensrcHdr!55}
\multicolumn{4}{c}{\textbf{\textit{AlloSpatial-4B}}} \\

\textit{Stage-1 (SFT)}
& 38.2 {\tiny \textcolor{red}{($\downarrow$6.9)}}
& 52.0 {\tiny \textcolor{green!60!black}{($\uparrow$23.7)}}
& -- \\

\textit{Stage-2 (RL)}
& \textbf{53.5} {\tiny \textcolor{green!60!black}{($\uparrow$8.4)}}
& \textbf{69.1} {\tiny \textcolor{green!60!black}{($\uparrow$40.8)}}
& \textbf{358} \\

\bottomrule
\end{tabular}
\end{adjustbox}
\end{minipage}

\vspace{-0.3cm}
\end{figure*}

\subsection{Ablation Studies and Additional Results}
\label{subsec:ablation}

\textbf{Effect of input-frame number.}
We study how AlloSpatial behaves under different observation budgets. Fig.~\ref{fig:frame} reports performance with $0$, $3$, $7$, $15$, and $24$ uniformly sampled input frames. AlloSpatial shows the clearest advantage when visual observations are sparse. In the $0$-frame setting, where the model relies only on structured AST text, AlloSpatial improves over Qwen3-VL by $+18.2$ points. With only $3$ and $7$ frames, it remains consistently stronger than both Qwen3-VL and GPT-5.2, reaching $50.0$ and $53.5$ overall, respectively. As the number of frames increases, the gap to Qwen3-VL narrows, suggesting that dense visual coverage can partially compensate for missing allocentric memory. This trend indicates that AlloSpatial is particularly effective under limited observations.

\textbf{Comparison with general thinking mode.}
We next compare AlloSpatial with the general thinking of Qwen3-VL-4B. As shown in Tab.~\ref{tab:ablation}, enabling thinking improves Qwen3-VL-4B on MindCube from $28.3$ to $36.1$, but brings only a marginal gain on VSI-Bench ($45.1\!\to\!45.5$) while increasing the average response length to $1064$ tokens. In contrast, AlloSpatial-4B after RL reaches $53.5$ on VSI-Bench and $69.1$ on MindCube with only average $358$ tokens. This suggests that the improvement does not come from longer chain-of-thought alone. Instead, the harnessed process provides a more efficient reasoning structure by grounding intermediate reasoning in explicit allocentric evidence.

\textbf{Comparison with QA-only training.} \label{sec:ablation}
We further perform the same two-stage training schedule on standard QA pairs constructed from the same data, to isolate whether the gains of AlloSpatial come from direct answer supervision or potential benchmark leakage. As shown in Tab.~\ref{tab:ablation}, QA-only SFT substantially improves MindCube from $28.3$ to $53.9$, but decreases VSI-Bench from $45.1$ to $43.1$. This suggests that answer supervision alone can fit certain benchmark patterns, but does not consistently improve spatial reasoning. QA-only RL slightly improves VSI-Bench to $46.2$ and maintains a strong MindCube score of $53.0$, yet remains well below AlloSpatial Stage-2, which reaches $53.5$ on VSI-Bench and $69.1$ on MindCube. This gap indicates that RL over QA supervision alone is insufficient; the main improvement comes from optimizing complete tool-use trajectories that follow the Spatial Reasoning Harness and interact with World2Mind during reasoning.

\section{Conclusion \& Limitations}

We presented \textbf{AlloSpatial}, an agentic framework that equips MFMs with allocentric spatial reasoning capability through the proposed World2Mind cognitive mapping sandbox and carefully designed spatial reasoning harness. Experiments on VSI-Bench and MindCube show that AlloSpatial improves proprietary models in a training-free setting and enables compact open-weight agents to outperform larger general-purpose and spatially specialized baselines. A remaining limitation of our study is numerical reasoning: current World2Mind representations provide robust allocentric structure, but reconstruction drift and imperfect metric calibration can still limit precise distance, size, and counting estimates. Future work may address this by incorporating stronger metric calibration and uncertainty-aware reconstruction into the cognitive mapping backend.

\newpage
{
    \small
    \bibliographystyle{plainnat}
    \bibliography{main}
}


\appendix

\section{Training \& Evaluation Configuration} \label{appendix:config}

\paragraph{RL training configuration.}
AlloSpatial-4B and AlloSpatial-8B are initialized from their corresponding supervised cold-start checkpoints at step 240, after three epochs of SFT. We optimize both agents with GSPO using \texttt{ms-swift} framework~\cite{zhao2024swiftascalablelightweightinfrastructure} and DeepSpeed ZeRO-2. The RL prompts are sampled from a shared training pool of 59,981 examples, including 49,981 VSI-style examples from VSI-590K and 10,000 examples from the MindCube training split. Inline validation is performed on 1,442 held-out examples, consisting of 392 VSI-Bench-tiny questions and 1,050 MindCube-tiny questions. For each prompt, the policy samples 8 rollouts, with at most 5 tool-interaction turns per rollout. The maximum sequence length is 32,768 tokens, and generated completions are capped at 8,192 tokens. The trajectory reward follows HGTR, combining structural compliance, answer accuracy, tool-use effectiveness, and length control with weights $0.15$, $0.60$, $0.10$, and $0.15$, respectively. Validation rewards are logged for monitoring but are not used for optimization.

\begin{table}[h]
\centering
\caption{\textbf{GSPO training configuration for the reported AlloSpatial checkpoints.}
The main paper reports the 600-step AlloSpatial-4B checkpoint and the 400-step AlloSpatial-8B checkpoint. Unique prompts are counted after grouping the 8 sampled rollouts per prompt.}
\label{tab:appendix_train_config}
\vspace{0.4em}
\begin{tabular}{lcc}
\toprule
\textbf{Setting} & \textbf{AlloSpatial-4B} & \textbf{AlloSpatial-8B} \\
\midrule
Initialization & Qwen3-VL-4B SFT step 240 & Qwen3-VL-8B SFT step 240 \\
Reported RL checkpoint & step 600 & step 400 \\
Trainer processes & 4 & 6 \\
Per-device train batch & 4 & 2 \\
Gradient accumulation & 4 & 4 \\
Generated trajectories/update & 64 & 48 \\
Unique prompts/update & 8 & 6 \\
Rollouts per prompt & 8 & 8 \\
Unique prompts to reported ckpt. & 4.8K & 2.4K \\
Learning rate & $1\times10^{-6}$ & $5\times10^{-7}$ \\
Precision / attention & bf16 / FlashAttention & bf16 / FlashAttention \\
Trainable modules & language modules & language modules \\
Optimizer infrastructure & full tuning + ZeRO-2 & full tuning + ZeRO-2 \\
Sampling temperature & 1.0 & 1.0 \\
KL coefficient $\beta$ & 0.01 & 0.0 \\
Eval / save interval & 100 steps & 50 steps \\
\bottomrule
\end{tabular}
\end{table}

Both runs use cosine learning-rate decay with a warmup ratio of $0.005$, freeze the vision tower and aligner modules, and adopt $\epsilon_{\mathrm{high}}=0.28$ with overlong-completion filtering. The training scripts are epoch-based rather than hard-coded with a fixed maximum number of steps; the reported checkpoints are selected by validation performance and inference efficiency. Detailed configuration information is shown in Tab.~\ref{tab:appendix_train_config}.

\paragraph{Evaluation configuration.}
For trained local agents, we evaluate AlloSpatial with lmms-eval framework~\cite{zhang2024lmmsevalrealitycheckevaluation}. The evaluator parses final responses from predefined answer tags, uses temperature $1.0$, allows up to 8 reasoning turns, and caps each completion at 8,192 new tokens. For VSI-Bench, the main post-trained evaluation follows the limited-observation setting with 7 uniformly sampled frames, while the frame-budget ablation evaluates $0$, $3$, $7$, $15$, and $24$ input frames. For training-free proprietary-model evaluation, we uniformly sample up to 32 frames to provide sufficient observations for external cognitive mapping. MindCube is evaluated using the provided multi-view images.

Across local-agent experiments, World2Mind uses the same default cognitive mapping pipeline, including monocular depth and pose estimation, SAM3-based open-vocabulary segmentation, confidence filtering, semantic point-cloud construction, AST serialization, and optional top-down map rendering. During evaluation, tool calls generated by the model are executed online, and the returned ASTs, route maps, or visualizations are inserted into the dialogue context for subsequent harness-guided reasoning.

\section{World2Mind Service Parallelization}
\label{appendix:World2Mind}

RL rollout generation can produce many concurrent World2Mind calls, while a single World2Mind process would serialize reconstruction and mapping requests. We therefore deploy World2Mind as a multi-process HTTP service during training and local evaluation. Each service process is assigned to an independent NPU worker and initializes the same cognitive mapping pipeline, including monocular geometry estimation, SAM3 segmentation, semantic point-cloud construction, AST generation, and route-map rendering. In our main configuration, eight World2Mind workers are launched in parallel to support concurrent tool execution.

Within each worker, NPU-intensive depth estimation and segmentation are executed under a worker-level lock, while downstream mapping, AST construction, and route-map generation proceed after NPU computation and are controlled by CPU-side concurrency limits. This separation prevents concurrent rollouts from over-subscribing NPU memory while allowing lightweight mapping operations to proceed efficiently.

Rollout and evaluation clients dispatch each \texttt{cognitive\_map} request to an available World2Mind worker using a simple load-balancing strategy based on current in-flight requests. Failed or unavailable workers are skipped and requests are retried on another endpoint when possible. This service design improves throughput and tail-latency stability under concurrent rollout generation, while keeping the World2Mind reconstruction pipeline and AST semantics unchanged.

\section{Spatial Reasoning Harness Prompts} \label{appendix:prompt}

The Spatial Reasoning Harness is implemented with three coupled components: a system message, a machine-readable tool interface, and a user-side reasoning protocol. For proprietary models, the tool interface is passed through function-calling schemas. For locally trained agents, the same interface is serialized as \texttt{<tool\_call>} blocks containing a JSON object with \texttt{name} and \texttt{arguments}. Visual tokens are placed before the reasoning protocol, ensuring that the model observes the frames or multi-view images before receiving step-by-step instructions.

\begin{promptbox}{System Message}
\textbf{Role.}
You are a spatial intelligence assistant that analyzes videos and images to answer spatial questions.

\medskip
\textbf{Available tools.}
\begin{itemize}[leftmargin=1.2em,itemsep=0.15em,topsep=0.2em]
    \item \texttt{world2mind}: builds an allocentric cognitive map and returns per-instance spatial data, including coordinates, sizes, and object relations.
    \item \texttt{view\_image}: renders cognitive-map visualizations, including top-down landmark maps, route maps, semantic maps, and point-cloud views.
\end{itemize}

\medskip
\textbf{Reliability note.}
The cognitive map is produced from monocular reconstruction and may contain missing objects, ghost instances, coordinate drift, or segmentation errors. Treat tool outputs as supplementary evidence rather than ground truth, and cross-validate them against direct visual observations before answering.

\medskip
\textbf{Answer format.}
Wrap the final answer in \texttt{<Answer></Answer>} tags.
\end{promptbox}

\begin{promptbox}{Tool Interface}

\textbf{\texttt{world2mind}.}
Generate a query-conditioned cognitive map from the current visual input. The tool automatically analyzes the video frames or images already provided in the conversation; the model does not provide file paths.

\medskip
\textbf{Returned knowledge.}
\begin{itemize}[leftmargin=1.2em,itemsep=0.15em,topsep=0.2em]
    \item \textbf{Landmark knowledge} (\texttt{knowledge\_type=landmark} or \texttt{both}): detected instances, metric coordinates, footprint size, and object relations.
    \item \textbf{Route knowledge} (\texttt{knowledge\_type=route} or \texttt{both}): traversable regions, camera trajectory, and grid-based route information.
\end{itemize}

\medskip
\begin{tabularx}{\linewidth}{@{}p{0.24\linewidth}p{0.22\linewidth}X@{}}
\toprule
\textbf{Parameter} & \textbf{Requirement} & \textbf{Meaning} \\
\midrule
\texttt{categories} 
& required; list of strings 
& Object categories to detect as landmarks, e.g., \texttt{chair}, \texttt{table}, \texttt{door}. \\

\texttt{scene\_type} 
& required; \texttt{indoor} or \texttt{outdoor} 
& Selects scene-specific reconstruction settings. \\

\texttt{knowledge\_type} 
& required; \texttt{landmark}, \texttt{route}, or \texttt{both} 
& Chooses landmark-only, route-only, or complete cognitive-map output. \\

\texttt{output\_format} 
& required; \texttt{rectangle} or \texttt{ellipse} 
& Specifies the landmark footprint representation. \\

\texttt{traversable\_categories} 
& required for \texttt{route} or \texttt{both} 
& Ground or surface categories used to infer passable regions, e.g., \texttt{floor}, \texttt{carpet}, \texttt{road}. \\
\bottomrule
\end{tabularx}

\medskip
\textbf{\texttt{view\_image}.}
Inspect a visualization generated by the most recent \texttt{world2mind} call. The required parameter \texttt{visualization\_type} must be selected from the available visualizations returned by \texttt{world2mind}, such as \texttt{landmark\_vis}, \texttt{route\_vis}, \texttt{pointcloud\_rgb\_topdown}, or \texttt{pointcloud\_semantic\_topdown}. The model should call \texttt{view\_image} when the YAML output is ambiguous, when object layout requires visual verification, or when map evidence must be checked against raw observations.
\end{promptbox}

\begin{promptbox}{Text-Form Tool Calls for Local Agents}
For local SFT/GSPO agents, tool calls are serialized as text while preserving the same schema:
\begin{lstlisting}[style=promptjson]
<tool_call>
{
  "name": "world2mind",
  "arguments": {
    "categories": ["chair", "table", "door"],
    "scene_type": "indoor",
    "knowledge_type": "both",
    "output_format": "rectangle",
    "traversable_categories": ["floor", "carpet"]
  }
}
</tool_call>
\end{lstlisting}

The schema requires \texttt{categories}, \texttt{scene\_type}, \texttt{knowledge\_type}, and \texttt{output\_format}. Malformed calls are returned as tool-error messages when applicable and are not treated as valid spatial evidence.
\end{promptbox}

\begin{promptbox}{User-Side Reasoning Protocol}
\textbf{Visual tokens.}
The video frames or multi-view images are placed before the textual instructions.

\medskip
\begin{description}[leftmargin=1.5em,itemsep=0.35em,topsep=0.2em]
    \item[\textbf{Step 1: Visual Clues.}]
    Describe concrete visual observations before any tool call, including visible objects, relative positions, spatial relations, and a preliminary answer when possible.

    \item[\textbf{Step 2.1: \texttt{world2mind} Tool Call.}]
    Call \texttt{world2mind} only when the question requires information that vision alone cannot reliably provide, such as metric distance, 3D coordinates, route layout, viewpoint transformation, or complex spatial relations.

    \item[\textbf{Step 2.2: Map Clues.}]
    If \texttt{world2mind} is called, summarize map evidence only: detected instances, coordinates, sizes, containment relations, route cells, trajectories, or camera orientations. Do not reconcile it with visual clues yet.

    \item[\textbf{Step 3.1: \texttt{view\_image} Tool Call.}]
    If the structured map is ambiguous or layout verification is needed, call \texttt{view\_image} using one of the available visualization types returned by \texttt{world2mind}.

    \item[\textbf{Step 3.2: Visualization Clues.}]
    Describe the rendered cognitive-map visualization, such as top-down layout, object arrangement, route trajectory, or conflicts with the raw visual input.

    \item[\textbf{Step 4: Cross-Validation.}]
    Compare visual evidence, map evidence, and visualization evidence. Identify conflicts caused by reconstruction drift, missing objects, false detections, or visual ambiguity, and decide which evidence should dominate.

    \item[\textbf{Step 5: Final Answer.}]
    Integrate the validated evidence and provide the conclusion. For single-word, numeric, or option questions, place only the final value inside the answer tag.
\end{description}

\medskip
\textbf{Question.} \texttt{\{query\}}
\end{promptbox}

\section{Case Analysis} \label{appendix:case}

\subsection{Case 1: Metric Closest-Point Reasoning}

\begin{promptbox}{User Turn: Input Observation and Query}
\textbf{Observation.} The input is a video of a kitchen. Eight approximately uniformly sampled frames are shown below as a two-row, four-column contact sheet.

\begin{center}
    \includegraphics[width=0.92\linewidth]{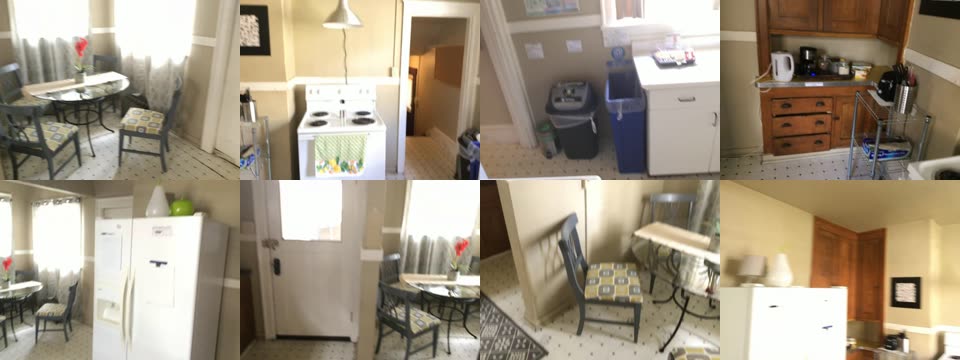}
\end{center}

\textbf{Question.} Determine the shortest distance between the towel and the refrigerator in meters. Please answer using a single word or phrase.
\end{promptbox}

\begin{promptbox}{Assistant Turn 1: Visual Clues and World2Mind Call}
\textbf{Visual clues.}
The model first observes a green patterned towel hanging on the oven door and a white refrigerator in other frames. Because the two objects are not visible in a single calibrated view, it estimates that direct visual inspection is insufficient for precise metric distance.

\medskip
\textbf{Tool call.}
\begin{lstlisting}[style=promptjson]
<tool_call>
{
  "name": "world2mind",
  "arguments": {
    "categories": ["towel", "refrigerator", "stove", "oven"],
    "knowledge_type": "landmark",
    "output_format": "rectangle",
    "scene_type": "indoor",
    "traversable_categories": ["floor"]
  }
}
</tool_call>
\end{lstlisting}
\end{promptbox}

\begin{promptbox}{Tool Result 1: World2Mind Landmark AST}
World2Mind returns a JSON tool result with \texttt{success=true}, scene id \texttt{scene0220\_01\_c70da29e-5e7}, available visualization \texttt{landmark\_vis}, and the following AST in the \texttt{landmark\_yaml} field.

\begin{lstlisting}[style=promptjson]
Map_Metadata:
  Coordinate_System: 2D Absolute Grid (meters)
  Origin_Ref: (-3.129, -1.293)
  AST_Format: Rectangle (center, width, height)
Spatial_Hierarchy:
- Node_ID: floor_01
  Spatial_Pose:
    Center: (-0.087, 1.449)
    Width: 4.117
    Height: 2.838
  Attributes:
    Height_Range: (1.179, 1.585)
    Area_m2: 7.1149
    Point_Count: 15383
- Node_ID: refrigerator_01
  Spatial_Pose:
    Center: (0.474, 1.475)
    Width: 0.924
    Height: 0.866
  Attributes:
    Height_Range: (-0.234, 1.525)
    Area_m2: 0.5815
    Point_Count: 2674
- Node_ID: stove_01
  Spatial_Pose:
    Center: (2.106, 0.691)
    Width: 0.65
    Height: 0.731
  Attributes:
    Height_Range: (0.554, 1.585)
    Area_m2: 0.4324
    Point_Count: 3588
  Contains_Children:
  - Node_ID: towel_01
    Spatial_Pose:
      Center: (1.918, 0.634)
      Width: 0.052
      Height: 0.304
    Attributes:
      Height_Range: (0.839, 1.207)
      Area_m2: 0.0102
      Point_Count: 1136
    Relation_To_Parent: Inside
\end{lstlisting}
\end{promptbox}

\begin{promptbox}{Assistant Turn 2: Map Calculation and Visualization Request}
\textbf{Map clues.}
The model reads \texttt{towel\_01} center $(1.918,0.634)$ m and \texttt{refrigerator\_01} center $(0.474,1.475)$ m. The center-to-center distance is about $1.671$ m, but the question asks for nearest-point distance. Using rectangle half-extents, the horizontal gap is
$1.444-(0.026+0.462)=0.956$ m and the vertical gap is
$0.841-(0.152+0.433)=0.256$ m. The nearest-boundary distance is therefore
$\sqrt{0.956^2+0.256^2}\approx0.990$ m.

\medskip
\textbf{Visualization request.}
The trajectory then requests a map visualization to verify the geometric layout before finalizing the answer.
\begin{lstlisting}[style=promptjson]
<tool_call>
{
  "name": "view_image",
  "arguments": {"visualization_type": "landmark_vis"}
}
</tool_call>
\end{lstlisting}
\end{promptbox}

\begin{promptbox}{Tool Result 2: Landmark Visualization}
The returned \texttt{landmark\_vis} image provides a top-down check of the refrigerator and stove/towel layout.

\begin{center}
    \includegraphics[width=0.66\linewidth]{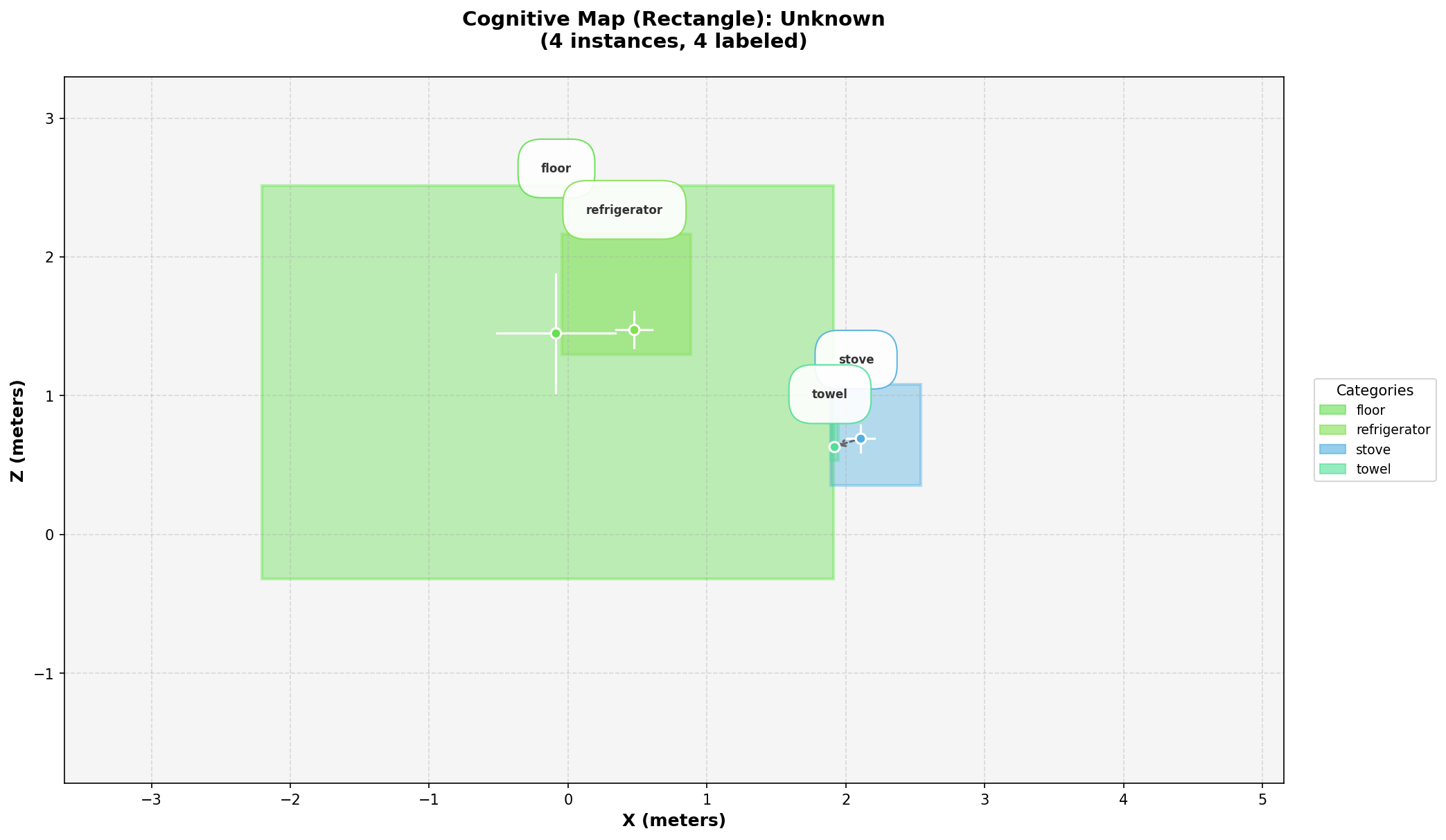}
\end{center}
\end{promptbox}

\begin{promptbox}{Assistant Turn 3: Cross-Validation, Final Answer, and Ground Truth}
\textbf{Visualization clues.}
The top-down landmark visualization places the refrigerator and stove/towel on different sides of the kitchen, with a gap consistent with the rectangle-derived estimate.

\medskip
\textbf{Cross-validation.}
The raw visual evidence suggests a short, plausible kitchen-scale separation, and the map does not introduce a visible contradiction. The agent therefore trusts the AST-derived nearest-point estimate.

\medskip
\textbf{Final answer.}
\begin{lstlisting}[style=promptjson]
Prediction: <Answer>0.99 m</Answer>
Ground truth (VSI): 1.1 m
Absolute error: 0.11 m
\end{lstlisting}
\end{promptbox}

\subsection{Case 2: Viewpoint-Conditioned Navigation}

\begin{promptbox}{User Turn: Input Observations and Query}
\textbf{Observation.} The input consists of four wall-aligned views of the same black chair: front, left, back, and right.

\begin{center}
\begin{minipage}{0.24\linewidth}
    \centering
    \includegraphics[width=\linewidth]{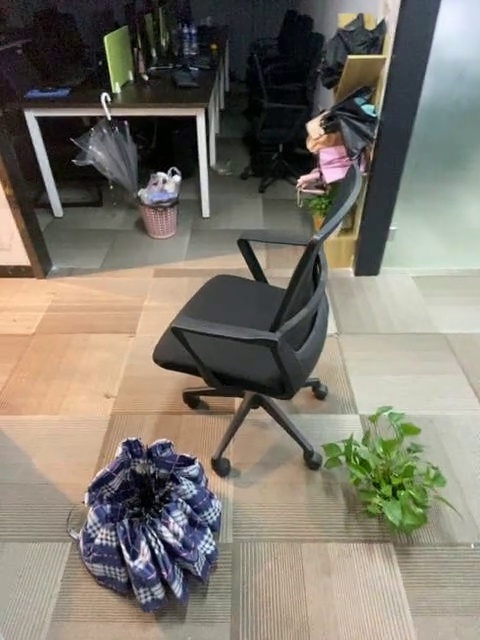}
    \centerline{\footnotesize Image 1}
\end{minipage}\hfill
\begin{minipage}{0.24\linewidth}
    \centering
    \includegraphics[width=\linewidth]{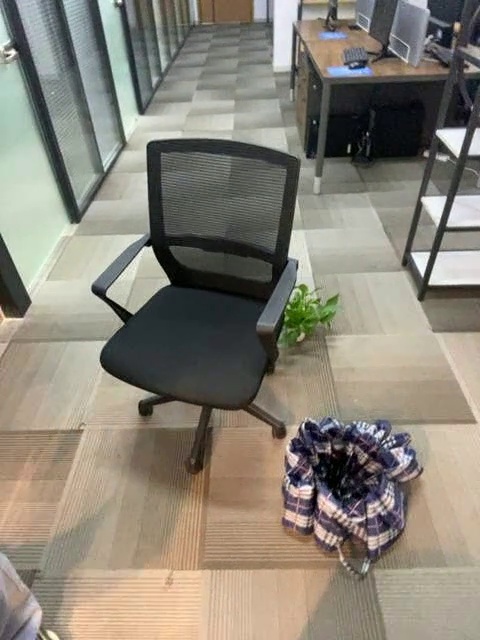}
    \centerline{\footnotesize Image 2}
\end{minipage}\hfill
\begin{minipage}{0.24\linewidth}
    \centering
    \includegraphics[width=\linewidth]{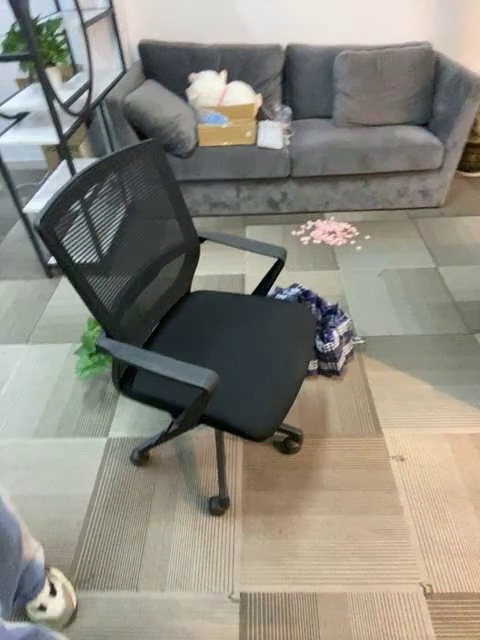}
    \centerline{\footnotesize Image 3}
\end{minipage}\hfill
\begin{minipage}{0.24\linewidth}
    \centering
    \includegraphics[width=\linewidth]{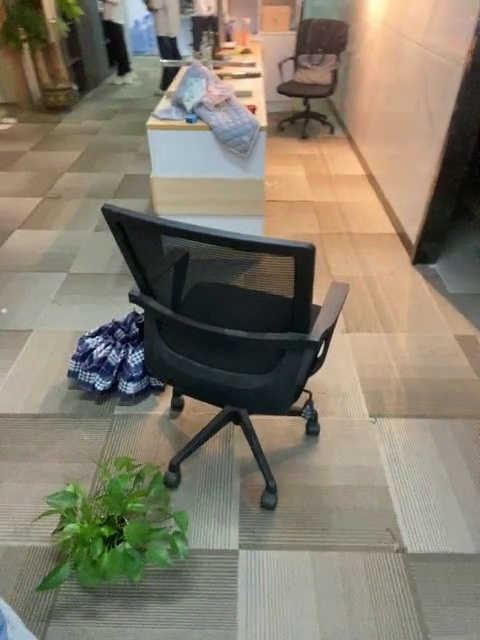}
    \centerline{\footnotesize Image 4}
\end{minipage}
\end{center}

\textbf{Question.} If the observer stands at the same spot and faces the same direction as image 4, then turns left and moves forward, will the observer get closer to the grey sofa? A. No B. Yes.
\end{promptbox}

\begin{promptbox}{Assistant Turn 1: Visual Clues and World2Mind Call}
\textbf{Visual clues.}
The model identifies image 1 as the front view of the chair, image 2 as the left view, image 3 as the back view, and image 4 as the right view. The grey sofa is clearly visible in image 3. From image 4, turning left appears to rotate the observer toward the direction where the sofa may lie, but the view transformation is ambiguous enough to benefit from an allocentric map.

\medskip
\textbf{Tool call.}
\begin{lstlisting}[style=promptjson]
<tool_call>
{
  "name": "world2mind",
  "arguments": {
    "categories": ["black chair", "grey sofa", "desk", "plant", "bag", "window"],
    "knowledge_type": "both",
    "output_format": "rectangle",
    "scene_type": "indoor",
    "traversable_categories": ["floor", "carpet", "tile flooring"]
  }
}
</tool_call>
\end{lstlisting}
\end{promptbox}

\begin{promptbox}{Tool Result 1: World2Mind AST and Route Knowledge}
World2Mind returns both landmark and route knowledge for scene \texttt{images\_4\_816b9d06-732}. The complete AST and route fields are shown below.

\textbf{AST.}
\begin{lstlisting}[style=promptjson]
Map_Metadata:
  Coordinate_System: 2D Absolute Grid (meters)
  Origin_Ref: (-2.281, -0.401)
  AST_Format: Rectangle (center, width, height)
  Camera_Views:
  - Frame: frame_000000
    Image: image 1
    Position: (-0.975, 0.668)
    Heading_Deg: 56.2
  - Frame: frame_000001
    Image: image 2
    Position: (0.921, 0.676)
    Heading_Deg: -57.6
  - Frame: frame_000002
    Image: image 3
    Position: (-0.0, 0.0)
    Heading_Deg: 0.3
  - Frame: frame_000003
    Image: image 4
    Position: (-0.17, 1.443)
    Heading_Deg: 161.8
Spatial_Hierarchy:
- Node_ID: grey sofa_01
  Spatial_Pose:
    Center: (1.524, 1.234)
    Width: 0.708
    Height: 1.022
  Attributes:
    Height_Range: (-0.654, 0.881)
    Area_m2: 0.4026
    Point_Count: 640
- Node_ID: black chair_01
  Spatial_Pose:
    Center: (-0.077, 1.314)
    Width: 0.532
    Height: 0.333
  Attributes:
    Height_Range: (-0.529, 0.49)
    Area_m2: 0.1173
    Point_Count: 2911
- Node_ID: bag_01
  Spatial_Pose:
    Center: (0.32, 1.278)
    Width: 0.412
    Height: 0.185
  Attributes:
    Height_Range: (0.311, 0.672)
    Area_m2: 0.0523
    Point_Count: 296
- Node_ID: plant_01
  Spatial_Pose:
    Center: (0.214, 1.863)
    Width: 0.307
    Height: 0.22
  Attributes:
    Height_Range: (-0.187, 0.221)
    Area_m2: 0.0419
    Point_Count: 516
\end{lstlisting}

\textbf{Route knowledge.}
\begin{lstlisting}[style=promptjson]
Route_Knowledge_Metadata:
  Coordinate_System: Grid (NxN cells)
  Grid_Divisions: 10
  Cell_Size: (0.474, 0.427)
  Scene_Bounds:
    Min: (-2.281, -0.401)
    Max: (2.458, 3.865)
  Total_Traversable_Cells: 32
  Estimated_Floor_Area_m2: 6.47
  Trajectory_Length: 4
  Note: Estimated_Floor_Area_m2 is the traversable floor area only. Actual room area
    approximately Floor_Area + furniture footprint area. Do NOT use Scene_Bounds to calculate
    room area -- Scene_Bounds is the axis-aligned bounding box of the entire 3D reconstruction
    and always significantly overestimates room size.
  Description: Route knowledge map showing traversable areas and camera movement path
  Scene: images_4_816b9d06-732
Traversable_Grid:
- (0, 3)
- (1, 3)
- (1, 4)
- (1, 6)
- (1, 7)
- (1, 8)
- (2, 3)
- (2, 6)
- (2, 7)
- (3, 2)
- (3, 3)
- (3, 4)
- (3, 5)
- (3, 6)
- (3, 7)
- (4, 2)
- (4, 3)
- (4, 4)
- (4, 5)
- (4, 6)
- (4, 7)
- (5, 3)
- (5, 4)
- (5, 5)
- (5, 6)
- (6, 3)
- (6, 4)
- (6, 5)
- (6, 6)
- (7, 3)
- (7, 5)
- (7, 6)
Camera_Trajectory: (2, 2) --> (2, 6) --> (0, 4) --> (4, 4)
Camera_Orientations:
- Frame: frame_000000
  Image: image 1
  Position: (-0.975, 0.668)
  Heading_Deg: 56.2
- Frame: frame_000001
  Image: image 2
  Position: (0.921, 0.676)
  Heading_Deg: -57.6
- Frame: frame_000002
  Image: image 3
  Position: (-0.0, 0.0)
  Heading_Deg: 0.3
- Frame: frame_000003
  Image: image 4
  Position: (-0.17, 1.443)
  Heading_Deg: 161.8
\end{lstlisting}
\end{promptbox}

\begin{promptbox}{Assistant Turn 2: Map Clues and View-Image Call}
\textbf{Map clues.}
The map places the grey sofa at $(1.524,1.234)$ and the black chair at $(-0.077,1.314)$. Image 4 has camera position $(-0.170,1.443)$ and heading $161.8^\circ$. Turning left changes the heading to approximately $71.8^\circ$, which points toward increasing $x$. Since the sofa has a much larger $x$ coordinate than the image-4 camera, moving forward after the left turn should reduce the distance to the sofa.

\medskip
\textbf{Tool call.}
\begin{lstlisting}[style=promptjson]
<tool_call>
{
  "name": "view_image",
  "arguments": {"visualization_type": "landmark_vis"}
}
</tool_call>
\end{lstlisting}
\end{promptbox}

\begin{promptbox}{Tool Result 2: Landmark Visualization}
The returned landmark visualization provides an explicit top-down check of the chair, sofa, and image-4 camera pose.

\begin{center}
    \includegraphics[width=0.68\linewidth]{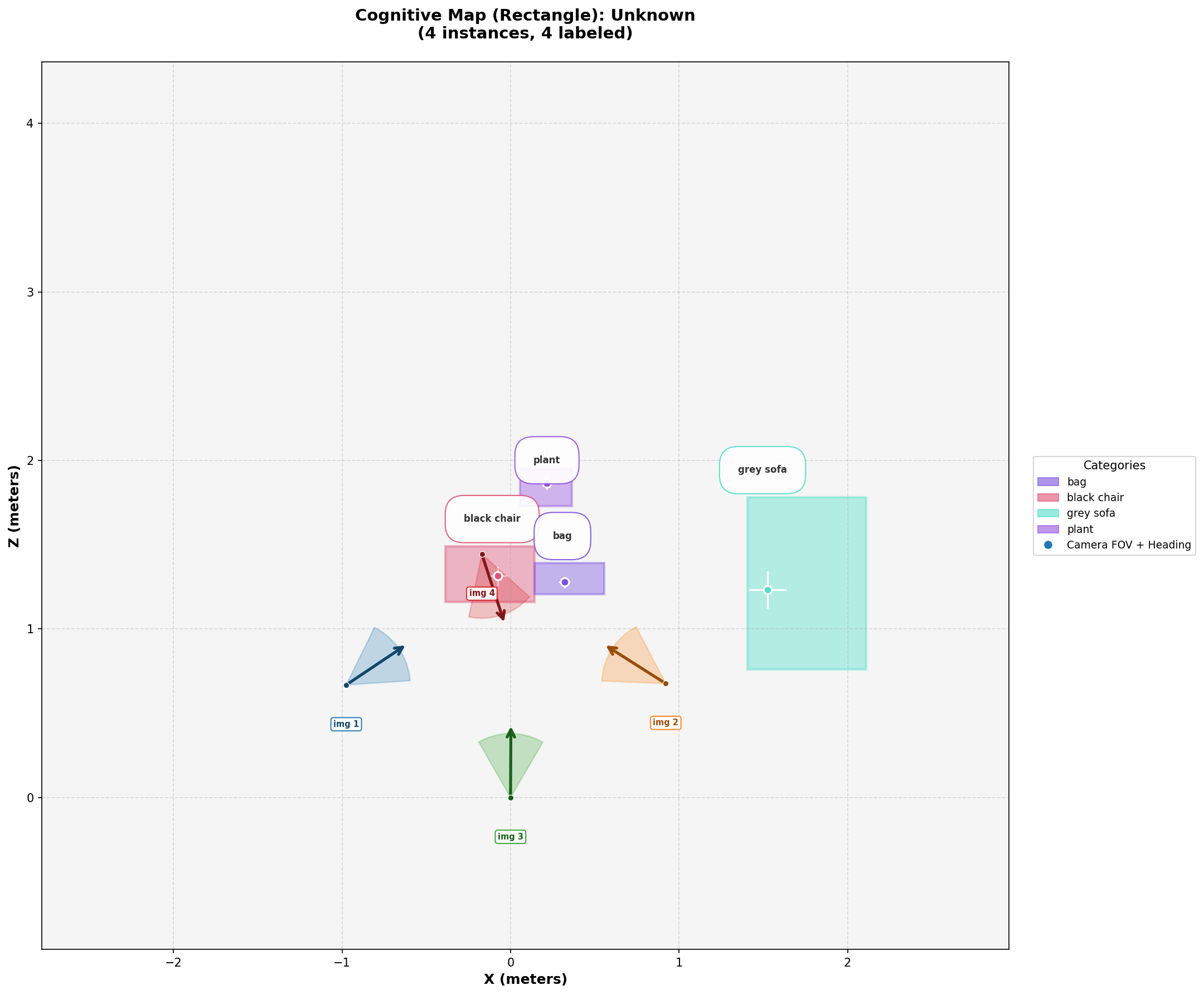}
\end{center}
\end{promptbox}

\begin{promptbox}{Assistant Turn 3: Visualization Clues, Cross-Validation, Final Answer, and Ground Truth}
\textbf{Visualization clues.}
The map shows the black chair near the center and the grey sofa to the right/east of the image-4 camera position. This agrees with the coordinate-level reading.

\medskip
\textbf{Cross-validation.}
Visual evidence from image 3 shows the sofa behind the chair, while image 4 is the right-side view. The map evidence shows that a left turn from heading $161.8^\circ$ points approximately toward heading $71.8^\circ$, which moves toward the sofa's larger $x$ coordinate. The visual and allocentric cues therefore agree.

\medskip
\textbf{Final answer.}
\begin{lstlisting}[style=promptjson]
Prediction: <Answer>B</Answer>
Ground truth (MindCube): B (Yes)
\end{lstlisting}
\end{promptbox}

\section{Computational Costs in Training and Inference}
\label{appendix:costs}

\paragraph{Training cost.}
In our setup, both AlloSpatial-4B and AlloSpatial-8B are trained on 8 HUAWEI Ascend 910B NPUs, taking approximately 60 and 40 NPU-hours, respectively, to reach the reported checkpoints. The 4B and 8B agents use 4 and 6 trainer processes, respectively. Around the reported checkpoints, one GSPO update typically takes 4--6 minutes, including multi-turn rollout generation, live World2Mind execution, reward computation, and policy optimization. The exact update time varies with the tool-use rate, the number of sampled World2Mind calls, and the reconstruction difficulty of each rollout batch.

\paragraph{Inference cost.}
Inference cost is determined primarily by the number of input frames and the number of World2Mind calls. Each AlloSpatial query may involve multiple reasoning turns, and each valid World2Mind call can trigger depth and pose estimation, segmentation, semantic alignment, AST construction, and optional route-map rendering. With the parallel World2Mind service, VSI-Bench-tiny evaluation takes roughly 12 minutes for 392 questions with concurrent evaluation and 8 World2Mind workers.




\newpage
\section*{NeurIPS Paper Checklist}

The checklist is designed to encourage best practices for responsible machine learning research, addressing issues of reproducibility, transparency, research ethics, and societal impact. Do not remove the checklist: {\bf The papers not including the checklist will be desk rejected.} The checklist should follow the references and follow the (optional) supplemental material.  The checklist does NOT count towards the page
limit. 

Please read the checklist guidelines carefully for information on how to answer these questions. For each question in the checklist:
\begin{itemize}
    \item You should answer \answerYes{}, \answerNo{}, or \answerNA{}.
    \item \answerNA{} means either that the question is Not Applicable for that particular paper or the relevant information is Not Available.
    \item Please provide a short (1--2 sentence) justification right after your answer (even for \answerNA). 
\end{itemize}

{\bf The checklist answers are an integral part of your paper submission.} They are visible to the reviewers, area chairs, senior area chairs, and ethics reviewers. You will also be asked to include it (after eventual revisions) with the final version of your paper, and its final version will be published with the paper.

The reviewers of your paper will be asked to use the checklist as one of the factors in their evaluation. While \answerYes{} is generally preferable to \answerNo{}, it is perfectly acceptable to answer \answerNo{} provided a proper justification is given (e.g., error bars are not reported because it would be too computationally expensive'' or ``we were unable to find the license for the dataset we used''). In general, answering \answerNo{} or \answerNA{} is not grounds for rejection. While the questions are phrased in a binary way, we acknowledge that the true answer is often more nuanced, so please just use your best judgment and write a justification to elaborate. All supporting evidence can appear either in the main paper or the supplemental material, provided in appendix. If you answer \answerYes{} to a question, in the justification please point to the section(s) where related material for the question can be found.

IMPORTANT, please:
\begin{itemize}
    \item {\bf Delete this instruction block, but keep the section heading ``NeurIPS Paper Checklist"},
    \item  {\bf Keep the checklist subsection headings, questions/answers and guidelines below.}
    \item {\bf Do not modify the questions and only use the provided macros for your answers}.
\end{itemize}


\begin{enumerate}

\item {\bf Claims}
    \item[] Question: Do the main claims made in the abstract and introduction accurately reflect the paper's contributions and scope?
    \item[] Answer: \answerYes{}
    \item[] Justification: The abstract and introduction state the paper's main contributions: AlloSpatial, World2Mind, the Spatial Reasoning Harness, and the RL-based internalization into open-weight agents. The claims are supported by evaluations on VSI-Bench and MindCube, including training-free proprietary-model results, trained Qwen3-VL-based agents, ablations, and limitations.

\item {\bf Limitations}
    \item[] Question: Does the paper discuss the limitations of the work performed by the authors?
    \item[] Answer: \answerYes{}
    \item[] Justification: The paper discusses limitations in the Conclusion and Limitations section, especially the remaining weakness on fine-grained numerical spatial reasoning due to reconstruction drift and imperfect metric calibration. The appendix also reports computational costs and implementation constraints of online World2Mind execution.

\item {\bf Theory assumptions and proofs}
    \item[] Question: For each theoretical result, does the paper provide the full set of assumptions and a complete (and correct) proof?
    \item[] Answer: \answerNA{}
    \item[] Justification: The paper does not present theoretical theorems or formal proofs.

    \item {\bf Experimental result reproducibility}
    \item[] Question: Does the paper fully disclose all the information needed to reproduce the main experimental results of the paper to the extent that it affects the main claims and/or conclusions of the paper (regardless of whether the code and data are provided or not)?
    \item[] Answer: \answerYes{}
    \item[] Justification: The paper describes the benchmarks, data splits, metrics, frame-sampling protocols, baselines, training stages, reward design, evaluation settings, prompt format, and World2Mind service configuration. Additional hyperparameters, computational costs, and case analyses are provided in the appendix.

\item {\bf Open access to data and code}
    \item[] Question: Does the paper provide open access to the data and code, with sufficient instructions to faithfully reproduce the main experimental results, as described in supplemental material?
    \item[] Answer: \answerNo{}
    \item[] Justification: The paper uses public benchmarks and datasets where available and provides detailed implementation, training, and evaluation settings in the appendix. Code and checkpoints are not included in the anonymous submission; we plan to release reproducibility materials subject to licensing and anonymization constraints.

\item {\bf Experimental setting/details}
    \item[] Question: Does the paper specify all the training and test details (e.g., data splits, hyperparameters, how they were chosen, type of optimizer) necessary to understand the results?
    \item[] Answer: \answerYes{}
    \item[] Justification: The main paper specifies datasets, benchmarks, metrics, frame budgets, baselines, model variants, and the two-stage training procedure. The appendix provides GSPO training configurations, reward weights, rollout settings, evaluation parameters, World2Mind parallelization, and computational costs.

\item {\bf Experiment statistical significance}
    \item[] Question: Does the paper report error bars suitably and correctly defined or other appropriate information about the statistical significance of the experiments?
    \item[] Answer: \answerNo{}
    \item[] Justification: The paper reports benchmark scores on official Tiny subsets and controlled ablations, but does not report error bars or confidence intervals. Repeating full RL training and proprietary-model evaluations multiple times would be computationally expensive; instead, the paper provides task-level breakdowns, ablations, and comparisons across multiple model families.


\item {\bf Experiments compute resources}
    \item[] Question: For each experiment, does the paper provide sufficient information on the computer resources (type of compute workers, memory, time of execution) needed to reproduce the experiments?
    \item[] Answer: \answerYes{}
    \item[] Justification: The appendix reports the training infrastructure, including 8 NVIDIA H200 GPUs, trainer process counts, GSPO update time, approximate wall-clock training time for AlloSpatial-4B and AlloSpatial-8B, World2Mind service workers, and inference throughput.

    
\item {\bf Code of ethics}
    \item[] Question: Does the research conducted in the paper conform, in every respect, with the NeurIPS Code of Ethics \url{https://neurips.cc/public/EthicsGuidelines}?
    \item[] Answer: \answerYes{}
    \item[] Justification: The work uses public benchmarks, existing foundation models, and generated tool-use trajectories for spatial reasoning research. It does not involve deception, human-subject experimentation, private personal data collection, or unsafe data release.


\item {\bf Broader impacts}
    \item[] Question: Does the paper discuss both potential positive societal impacts and negative societal impacts of the work performed?
    \item[] Answer: \answerYes{}
    \item[] Justification: The paper discusses AlloSpatial as foundational research toward more spatially capable multimodal and embodied agents. It also notes that incorrect spatial reasoning or over-trusting reconstructed maps can be harmful in safety-critical settings, motivating the harness-based cross-validation design and the limitation discussion on metric reliability.
    
\item {\bf Safeguards}
    \item[] Question: Does the paper describe safeguards that have been put in place for responsible release of data or models that have a high risk for misuse (e.g., pre-trained language models, image generators, or scraped datasets)?
    \item[] Answer: \answerNA{}
    \item[] Justification: The paper does not introduce a scraped dataset, image generator, or high-risk pretrained foundation model release. The proposed method is evaluated as a spatial reasoning framework, and any future release of code or checkpoints will follow the licenses and usage restrictions of the underlying models and datasets.

\item {\bf Licenses for existing assets}
    \item[] Question: Are the creators or original owners of assets (e.g., code, data, models), used in the paper, properly credited and are the license and terms of use explicitly mentioned and properly respected?
     \item[] Answer: \answerYes{}
    \item[] Justification: The paper cites the existing datasets, benchmarks, foundation models, geometry models, segmentation models, and spatial reasoning baselines used in the study. We use these assets for research evaluation and training under their respective terms and do not redistribute restricted proprietary models or datasets.


\item {\bf New assets}
    \item[] Question: Are new assets introduced in the paper well documented and is the documentation provided alongside the assets?
    \item[] Answer: \answerYes{}
    \item[] Justification: The paper introduces AlloSpatial, World2Mind, the Spatial Reasoning Harness, and trained AlloSpatial agents. The method, prompts, tool schema, reward design, training configuration, evaluation setup, and service parallelization are documented in the main paper and appendix; release of code or checkpoints will follow anonymization and licensing constraints.


\item {\bf Crowdsourcing and research with human subjects}
    \item[] Question: For crowdsourcing experiments and research with human subjects, does the paper include the full text of instructions given to participants and screenshots, if applicable, as well as details about compensation (if any)? 
    \item[] Answer: \answerNA{}
    \item[] Justification: The paper does not involve crowdsourcing experiments or research with human subjects. All evaluations are performed on existing benchmarks and generated model trajectories.


\item {\bf Institutional review board (IRB) approvals or equivalent for research with human subjects}
    \item[] Question: Does the paper describe potential risks incurred by study participants, whether such risks were disclosed to the subjects, and whether Institutional Review Board (IRB) approvals (or an equivalent approval/review based on the requirements of your country or institution) were obtained?
    \item[] Answer: \answerNA{}
    \item[] Justification: The paper does not involve human-subject research, user studies, or collection of participant data. Therefore, IRB approval or equivalent review is not applicable.


\item {\bf Declaration of LLM usage}
    \item[] Question: Does the paper describe the usage of LLMs if it is an important, original, or non-standard component of the core methods in this research? Note that if the LLM is used only for writing, editing, or formatting purposes and does \emph{not} impact the core methodology, scientific rigor, or originality of the research, declaration is not required.
 \item[] Answer: \answerYes{}
    \item[] Justification: LLMs and MFMs are core components of the research. The paper describes the use of proprietary models for training-free evaluation and trajectory distillation, Qwen3-VL as the open-weight backbone for AlloSpatial agents, and LLM-based multi-turn tool-use reasoning as part of the proposed method.

\end{enumerate}

\end{document}